%% file: main.tex
\documentclass[10pt,twocolumn,letterpaper]{article}
\usepackage[T1]{fontenc}

\usepackage[pagenumbers]{cvpr}

\input{preamble}

\definecolor{cvprblue}{rgb}{0.21,0.49,0.74}
\usepackage[pagebackref,breaklinks,colorlinks,allcolors=cvprblue]{hyperref}

\def\confName{CVPR}
\def\confYear{2027}

\title{Per-Light Reshading Learned from Single Images}

\author{Kuan Sheng Chen\\
University of Illinois Urbana-Champaign\\
{\tt\small kc81@illinois.edu}
\and
Svetlana Lazebnik\\
University of Illinois Urbana-Champaign\\
{\tt\small slazebni@illinois.edu}
\and
David Forsyth\\
University of Illinois Urbana-Champaign\\
{\tt\small daf@illinois.edu}
}

\def\TEASERONTOP{}
\begin{document}
\maketitle
\input{teaser_strip}
\input{abstract}
\input{intro}
\input{related}
\input{method}
\input{experiments}
\input{conclusion}
\clearpage
\section*{Use of Large Language Models}
Large language models (LLMs) were used as assistive tools in this work.
They assisted with code design, with running and organising experiments, and with cleaning code.
LLMs were also used to assist in writing this paper and in searching for and checking related work and references.
All LLM-assisted code, results and text were reviewed and verified by the authors, who take full responsibility for the
content of this paper.
{
    \small
    \bibliographystyle{ieeenat_fullname}
    \bibliography{refs}
}

\input{X_suppl}

\end{document}

%% file: preamble.tex
\usepackage{cuted}
\usepackage{currfile}
\usepackage{bm}
\usepackage{pifont}
\providecommand{\cmark}{\ding{51}}\providecommand{\xmark}{\ding{55}}
\AtEndPreamble{\crefname{appendix}{Sec.}{Secs.}\Crefname{appendix}{Section}{Sections}\crefname{subappendix}{Sec.}{Secs.}\Crefname{subappendix}{Section}{Sections}}

%% file: teaser_strip.tex
\begin{strip}
  \centering
  \IfFileExists{figs/teaser_arxiv.pdf}{\includegraphics[width=\linewidth]{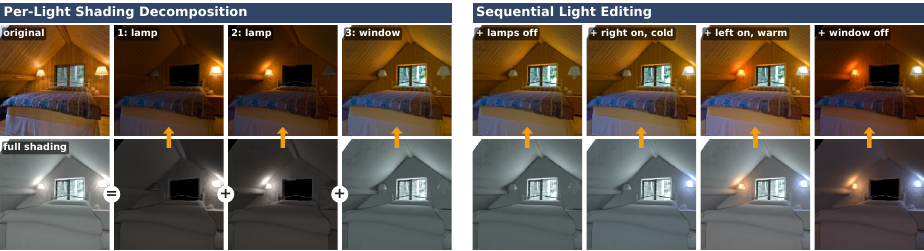}}{\fbox{\parbox[c][4.2cm][c]{0.95\linewidth}{\centering Teaser}}}
  \captionsetup{hypcap=false}
  \captionof{figure}{\textbf{PLRS (per-light reshading from single images) enables real-time light editing from one photograph.} Our method splits the photograph's shading into one layer per lamp and per window (left); above each layer is the scene lit by that light alone. Each light can then be switched, dimmed and recoloured, and once the photograph is decomposed every edit takes about $0.1$\,ms. Edits can be chained (right). Each edit is rendered from the albedo and the edited layers, above its shading.}
  \captionsetup{hypcap=true}
  \label{fig:teaser}
\end{strip}

%% file: abstract.tex
\begin{abstract}
To relight a room, you act on lights: switch on a ceiling light, dim a lamp and so on.
Relighting a photograph should be the same.
Given a single still image, we propose a model that produces a per-light decomposition: it recovers the parameters of each luminaire
(light-emitting object, windows included) visible in the scene.
Each luminaire has an associated shading layer, computed using physical considerations from the parameters.
Editing the parameters --- changing the colour of a light, say --- yields a corresponding edit to the shading layer and so the
image, in about 0.1 ms, roughly 20,000 times faster than the fastest method that reports a time.
The model is learned from single images of real scenes.
Our training uses no rendered scenes, multi-illumination captures or per-light labels, so our model applies to a situation where an
agent just observes the world in which it moves.
We built a small multi-illumination benchmark of real pictures of real scenes.
Extensive experimental comparison using both current image metrics and human preferences reveals that our method outperforms
alternatives trained on single-illumination data.
\end{abstract}

%% file: intro.tex
\section{Introduction}
\label{sec:intro}

\ifdefined\TEASERONTOP\else\input{teaser}\fi

An agent --- perhaps even a person --- moves through a complicated world in which the lighting changes all the time.
How can it build a model of how appearance changes with lighting?
There is good evidence that multi-light data, many pictures of the same scene under different lights, yields good
models~\cite{li2022physically,lightlab,liang2026luxremix}.
But such data is not natural: to obtain it, the agent would have to wait in one spot for the sun to move, match pictures to
pictures it took in the past, or turn switches on and off.
Could the agent learn such a model from \emph{single images} alone: ordinary photographs, each under whatever lighting
it happens to have?
There is reason to expect it could: albedo predictors can be learned without ever collecting the true albedo of real
scenes~\cite{forsyth2022paradigms}.

This paper shows that the answer is yes.
Our model combines an existing intrinsic image predictor~\cite{careaga2024colorful}, which gives a single shading for the scene,
with physical modelling to relight a photograph through explicit parameters of its visible \emph{luminaires}, the objects in it
that emit rather than reflect light.
Generative relighters represent what the sources do to the image; we represent the sources themselves.
A luminaire is a handful of numbers: whether it is on, its colour, its intensity, where it sits and how it throws light, so
the agent can ask how the room would look if this lamp turned orange or that one went off (\cref{fig:teaser}).
Windows are luminaires too: where recent relighters~\cite{lightlab,liang2026luxremix} treat daylight as ambient light, our model can dim, switch off or recolour each window's daylight (\cref{fig:window-edit}).
Each luminaire's direct light is computed physically from the scene's geometry; learned components approximate interreflections and the luminaires' own appearance.
Because each luminaire can light only what it can physically reach, supervising the total shading of a photograph constrains
how it splits among its luminaires.

A satisfactory model produces an image of exactly the same scene, lit exactly as requested: asked how the room
looks with this light switched off, it should produce the real photograph of the room in that condition.
Existing models aim at this too, but differ from ours in what they are trained on or given, so none is fully comparable, though many are partially (\cref{tab:intro-methods}).
Li \etal~\cite{li2022physically} is the most comparable in what it does, since it too edits the photograph's own lamps and windows through an
explicit shading, but it is trained on rendered multi-light rooms, with the shading of each light as ground truth; the others need the
user to scribble the change~\cite{scribblelight} or an image of the intended lighting~\cite{xing2025luminet}, or are
trained on undescribed~\cite{wu2025qwenimage} or multi-light data~\cite{lightlab,liang2026luxremix}.
We nonetheless compare with all of them.
Models trained on multi-light data show its benefit and serve as a reference rather than a target: we aim for a model that an agent can learn only from what it happens to see.

\input{tab_intro_methods}

To measure edits against reality, we captured indoor scenes under different combinations of their own lights, giving pairs of real photographs that differ by switched lights (\cref{fig:benchmark}).
In comparisons on these pairs and two public benchmarks, the best models trained on multi-light data lead on their own data, but lacking such data does not make relighting impossible.
Qwen-Image-Edit hallucinates, removes or replaces a light fixture in 73\% of pairs, changing the scene rather than its lighting. Ours keeps every fixture and outperforms ScribbleLight, which is also trained on single photographs but alters fixtures in 30\% of pairs; on these pairs and LuxRemix's renders, ours also has lower LPIPS than Li~\etal, trained on multi-light renders.
Image metrics are not a reliable guide to relighting quality: under LPIPS~\cite{zhang2018unreasonable}, the unedited photograph is closer to the target than most edits by most models, yet people strongly prefer our edits both to it and to the edits of ScribbleLight and Li~\etal (\cref{sec:comparisons}).
Where metrics are unreliable, fluid interaction has real advantages: our $12.8$\,M-parameter network decomposes a photograph once, in about $2$\,s, then each edit takes about $0.1$\,ms, roughly $20{,}000\times$ faster than the fastest method that reports a time, so edits run at $10$\,kHz (\cref{tab:intro-methods}).

Our contributions are:
\begin{itemize}
  \item per-light shading learned without per-light labels: because each luminaire's direct light is computed physically,
  supervising a photograph's total shading constrains how it splits among the luminaires;
  \item to our knowledge, the first model that relights a photograph by editing its own luminaires, including windows,
  trained only on single images, with real-time edits;
  \item a benchmark dataset of real indoor photographs under combinations of each room's own lights, with labelled light
  states and masks, to be released; to our knowledge, no such dataset is public.
\end{itemize}

%% file: teaser.tex
\begin{figure*}[t]
  \centering
  \IfFileExists{figs/teaser.pdf}{\includegraphics[width=\linewidth]{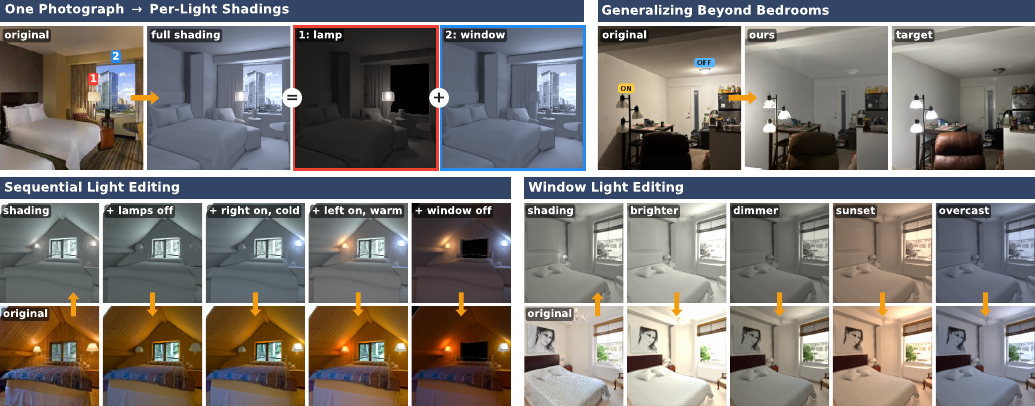}}{\fbox{\parbox[c][4.2cm][c]{0.95\linewidth}{\centering Teaser}}}
  \caption{\textbf{ShadeSplit enables real-time light editing from a single photograph.} Our method splits the photograph's shading into one layer per lamp and per window, which add up to the full shading (top left; the lamp $\times3$ for display). Each light can then be switched, dimmed and recoloured, and once the photograph is decomposed every edit takes about $0.1$\,ms. Learned only from LSUN bedroom photographs, it carries over to other indoor scenes: our edit of a real kitchen pair next to the target (top right). Edits can be chained (bottom left: both lamps off, the right one back on cold and the left warm, then the window off), and daylight is a light like any other (bottom right: the window brightened, dimmed and turned to sunset and overcast light). Each edit is rendered from the albedo and the edited layers, below its shading.}
  \label{fig:teaser}
\end{figure*}

%% file: tab_intro_methods.tex
\providecommand{\hcell}[1]{\begin{tabular}[b]{@{}c@{}}#1\end{tabular}}
\begin{table}[t]
\caption{\textbf{The compared methods.} \cmark{} marks what a method does or avoids; --: not reported; $^\dagger$sees the target photograph. \emph{Params}: networks run for an edit ($\geq$: backbone only). \emph{Time / edit}: $^*$as reported; ours after about $2$\,s per photograph.}
\label{tab:intro-methods}
\centering
\footnotesize
\setlength{\tabcolsep}{3pt}
\resizebox{\linewidth}{!}{%
\begin{tabular}{@{}lccccrr@{}}
\toprule
 & & \multicolumn{2}{c}{Per-light edit} & & & \\
\cmidrule(lr){3-4}
Method & \hcell{Trained on single\\photographs} & lamp & window & \hcell{Explicit\\shading} & Params & \hcell{Time\\/ edit} \\
\midrule
Ours                                   & \cmark & \cmark & \cmark & \cmark & 12.8\,M & 0.1\,ms \\
\midrule
ScribbleLight~\cite{scribblelight}     & \cmark & \xmark & \xmark & \xmark & 1.7\,B  & 3.9\,s \\
Li \etal~\cite{li2022physically}       & \xmark & \cmark & \cmark & \cmark & 139\,M  & 1.9\,s$^*$ \\
LuxRemix~\cite{liang2026luxremix}      & \xmark & \cmark & \xmark & \xmark & $\geq$12\,B & -- \\
LightLab~\cite{lightlab}               & \xmark & \cmark & \xmark & \xmark & $\geq$2.6\,B & 5\,s$^*$ \\
LumiNet$^\dagger$~\cite{xing2025luminet} & \xmark & \xmark & \xmark & \xmark & 2.1\,B  & 3.9\,s \\
Qwen-Image-Edit~\cite{wu2025qwenimage} & --     & \xmark & \xmark & \xmark & 28.9\,B & 81\,s \\
\bottomrule
\end{tabular}}
\end{table}

%% file: related.tex
\section{Related Work}
\label{sec:related}

\paragraph{Intrinsic images.}
Intrinsic decomposition separates a photograph into reflectance and
shading~\cite{land1971retinex,barron2015sirfs,li2018cgintrinsics,li2018bigtime}; recent methods predict grey~\cite{careaga2023ordinal} or coloured diffuse shading~\cite{careaga2024colorful},
or build on diffusion priors~\cite{zeng2024rgbx,ke2025marigold,luo2024intrinsicdiffusion},
and are evaluated on real photographs with reflectance judgements~\cite{bell2014iiw} and shading
annotations~\cite{kovacs2017saw,li2018cgintrinsics}.
All output a single shading map, which can be rescaled but not relit per source; we distil such a map~\cite{careaga2024colorful}
into a sum of per-light layers.

\paragraph{Lights in a single photograph.}
Per-light layers have required captured stacks~\cite{debevec2000reflectance,boyadzhiev2013user}, flash/no-flash pairs~\cite{petschnigg2004flash}
or illuminants of distinct colour~\cite{hsu2008light,hui2018illuminant,hui2019separate,kim2021lsmi}; from one photograph, they have been separated
by flash/ambient models~\cite{maralan2023flash} or per-image optimisation~\cite{zhang2017illumination}.
Lighting estimators~\cite{gardner2017learning,dastjerdi2023everlight,phongthawee2024diffusionlight,garon2019fast,li2020inverse,wang2021learning,srinivasan2020lighthouse,gardner2019deep,weber2022editable}
mostly serve object insertion~\cite{debevec1998rendering,karsch2011rendering} or rescale detected sources~\cite{karsch2014automatic}.
In contrast, we tie each light to a visible lamp, ceiling fixture or window as a 3D emitter, and learn the split from real photographs
instead of optimising it per image.

\paragraph{Editing the light of a real room from one image.}
Closest to us, Li~\etal~\cite{li2022physically} reconstruct a room's lamps and windows from one image and render each light's
direct shading physically with ray-traced shadows, as we do, but learn from rendered rooms with each light's shading as ground truth~\cite{li2021openrooms};
we learn the split among lights from single real photographs, without per-light supervision.
Careaga and Aksoy~\cite{careaga2025physically} train on unpaired real photographs but light the scene with user-placed lights
rather than its own sources.
Other methods edit lights through GAN pseudo-pairs or latents~\cite{cui2022local,bhattad2024stylitgan}, transfer a reference
image's lighting~\cite{zhang2024latent}, learn a residual renderer~\cite{sengupta2019neural}, render user-authored
lights~\cite{farinha2026pixlrelight}, or relight multi-view, RGB-D or outdoor
captures~\cite{zhang2016emptying,wu2023fipt,zhu2023i2sdf,lin2025iris,philip2021freeviewpoint,philip2019multiview,yu2020selfsupervised,liu2020factorize,tzevelekakis2025solo}.
Concurrent with our work, FreeLit~\cite{yen2026freelit} uses our data regime but leaves shadows and interreflection to a
diffusion model, and has no window source.
In contrast, we recover the photograph's own lamps, ceiling fixtures and windows, each with the physical model of its kind, and train on real photographs.

\paragraph{Generative relighting and light-source control.}
Diffusion relighters mostly control light globally~\cite{iclight,liang2025diffusionrenderer,he2025unirelight,xing2025luminet,kocsis2024lightit,zeng2024rgbx,zeng2024dilightnet}.
Others target individual lights through scribbles~\cite{scribblelight}, captured on/off pairs~\cite{lightlab}, one-light-at-a-time
layers learned from synthetic scenes~\cite{liang2026luxremix} or per-light
tokens~\cite{chaturvedi2026tokenlight,zhou2026lightmover}, and general-purpose image editors now also switch a lamp placed in a
scene~\cite{kuechler2026do}.
These models synthesise every pixel with priors of one to tens of billions of parameters, their results change with the seed, and
daylight is folded into an ambient term~\cite{lightlab,liang2026luxremix}.
In contrast, we output deterministic, explicit per-light shading over a fixed albedo, with windows as sources in their own right; unlike
LuxRemix's per-light images, our layers are shading fields computed from physical light models rather than generated.

\paragraph{Learning with and without multi-light images.}
Most methods that control individual lights see the same room under different lights in training: rendered one light at a
time~\cite{li2021openrooms,li2022physically,liang2026luxremix}, photographed~\cite{lightlab,murmann2019miiw,li2018bigtime,farinha2026pixlrelight}
or generated by a GAN~\cite{cui2022local,xing2025luminet}.
Real images of this kind are scarce: photographs of a room's own lights switched~\cite{cui2022local,lightlab,chaturvedi2026tokenlight}
are not public, and public captures switch a lamp placed in the scene~\cite{kuechler2026do}, photograph each light alone for
multi-view relighting~\cite{serranolozano2026synclight}, or target white balance~\cite{kim2021lsmi}.
Few methods learn from single photographs~\cite{careaga2025physically,scribblelight,yen2026freelit}; of these, ours is the only one
that edits the photograph's own lights as physical emitters, each with its direct light computed from a physical model.
A luminaire detector is easy to build; given one, a useful per-luminaire shader can be learned without multi-light
images.

%% file: method.tex
\newcommand{\Lset}{\mathcal{L}}
\newcommand{\Wset}{\mathcal{W}}
\newcommand{\pos}[1]{\left[#1\right]_{+}}
\newcommand{\avg}[1]{\left\langle #1\right\rangle}
\newcommand{\SL}{\operatorname{SL}}

\section{Method}
\label{sec:method}

\begin{figure*}[t]
  \centering
  \IfFileExists{figs/overview.pdf}{\includegraphics[width=\linewidth]{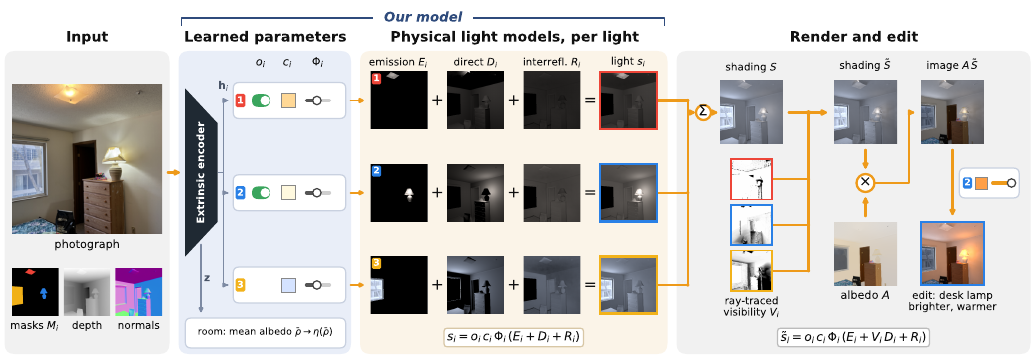}}{\fbox{\parbox[c][4cm][c]{0.9\linewidth}{\centering Overview figure to be drawn}}}
  \caption{\textbf{Overview.} From a photograph we estimate depth, normals and light masks. Each light $i$ has explicit parameters (on/off $o_i$, colour $c_i$, intensity $\Phi_i$; other parameters in \cref{sec:lamps}) and a shading layer $s_i$: its emission $E_i$, direct light $D_i$ and interreflected light $R_i$. The layers sum to the shading $S$; with ray-cast visibility $V_i$ on each $D_i$ they give $\tilde S$, the image is a fixed albedo times $\tilde S$, and an edit changes a light's parameters.}
  \label{fig:overview}
\end{figure*}
\begin{figure*}[t]
  \centering
  \includegraphics[width=0.76\linewidth]{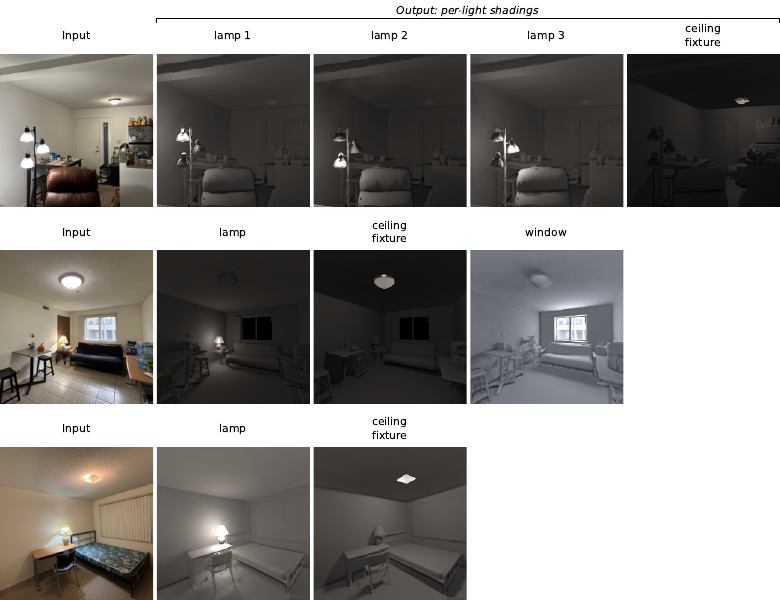}
  \caption{\textbf{Per-light shading decomposition.} Each input is followed by the
  shading layer our model gives each of its lights that is on; the layers add up to the input's shading, and an edit changes a light's parameters. More results in the appendix (\cref{fig:decomp-gallery}).}
  \label{fig:decomp-ours}
\end{figure*}

\Cref{fig:overview} gives an overview of our model.
Given a single photograph $\mathbf{I}$ of an indoor scene, we estimate an editable diffuse shading for each visible light source;
these per-light shadings sum to the scene's shading (\cref{fig:decomp-ours}; more in the appendix, \cref{fig:decomp-gallery}), and editing them changes its illumination.
In daylit rooms the window's layer dominates, so on one shared brightness scale lamp layers would be nearly black;
our figures therefore show lamp layers $\times3$ where a window lets in daylight and $\times1$ otherwise (in \cref{fig:teaser}, also within the full shading, so that the layers still add up to it).
Instead of regressing them directly, our network predicts the parameters of physics-based light models (\cref{fig:overview}).
An edit changes these parameters; the edited image is the albedo times the edited shading, with shadows ray-cast against the
scene's estimated geometry.
The albedo comes from a pretrained intrinsic model, filtered to remove the shading it retains (appendix, Sec.~\ref{supp:albedo}).
Windows are light sources in their own right, not part of an ambient term, and each source's bounced light is tied to its own
direct light: whenever a visible light is on, everything that lights the room belongs to some source.

\subsection{Shading formation and inputs}
\label{sec:scene}

\paragraph{Shading formation.}
\label{sec:formation}
\label{sec:fallback}
Our shading model has two cases, depending on whether the scene has a visible light source that is on:
\begin{equation}
  S_x = \begin{cases}
    s^{\mathrm{scene}}_{x}, & \text{no visible light on},\\
    \sum_{i\in\Lset\cup\Wset} s_{i,x}, & \text{otherwise},
  \end{cases}
  \label{eq:cast}
\end{equation}
\begin{equation}
  s_{i,x} = o_i\, c_i\, \Phi_i \big( E_{i,x} + D_{i,x} + R_{i,x} \big).
  \label{eq:formation}
\end{equation}
\emph{No visible light on} (every visible lamp predicted off and no transparent window) means the room is lit from outside the
frame; we then predict a single shading $s^{\mathrm{scene}}$ for the whole scene, as an intrinsic decomposition gives, which cannot be edited light by light (appendix, Sec.~\ref{supp:fallback}).

\emph{Otherwise}, we write the shading as a sum over the per-light shadings of visible lamps $\Lset$ and windows $\Wset$.
Here $o_i\in\{0,1\}$ switches light $i$ on or off, $c_i\in\mathbb{R}^3_{+}$ is its colour
and $\Phi_i$ its intensity. $E_i$ is its emission, zero outside its own mask.
$D_i$ and $R_i$ are the direct and indirect (interreflected) light it casts (\cref{sec:lamps,sec:windows,sec:interreflection}).
$D_i$ and $R_i$ are zero on window pixels, which are not indoor surfaces.

\paragraph{Inputs.}
Each visible light source $i$ has a mask $M_i$ (for a window, its transparent part) and a type, lamp (\cref{sec:lamps}) or
window (\cref{sec:windows}); masks come from a text-conditioned segmentation model (details in the appendix,
Sec.~\ref{supp:masks}) and, at inference, can also be drawn by hand.
A monocular estimator gives depth and surface normals $\mathbf{n}_x$ (\cref{sec:training}), and depth gives each pixel's 3D
position $\mathbf{X}_x$ through a fixed camera calibration, one for LSUN and the lens's own for our paired photographs (appendix,
Sec.~\ref{supp:geometry}).

\subsection{Physical models}
\label{sec:lightmodels}

Our shading model is built on illumination physics. There are two kinds of luminaires: lamps and windows. These are parametric physical models, while interreflection is a physical approximation whose parameters are estimated from the image (\cref{sec:learned}).

\subsubsection{Lamps}
\label{sec:lamps}

\paragraph{Parameters.} From each lamp's feature $\mathbf{h}_i$, RoI-pooled from the photo and normals by an extrinsic encoder (\cref{sec:training}), small heads predict its
on/off state $o_i$, colour $c_i$, shading intensity $\Phi_i$, emission gain $\gamma_i$ (appendix, \cref{eq:lamp-params}), position offset and lobe weights, all used below.
The lamp's emission, how it appears in the image, is $E_{i,x}=M_{i,x}\,\gamma_i\,\hat\xi_x$, zero outside its mask, where $\hat\xi$ is the fixture's emission pattern (a bright shade, a dark pole), predicted by a small CNN from the photo and normals.

\paragraph{Direct light.}
We model a lamp as a point emitter with an angular lobe,
\begin{equation}
  D_{i,x} = f_i(\boldsymbol{\omega}_{i,x})\,
  \frac{\pos{\mathbf{n}_x\!\cdot(-\boldsymbol{\omega}_{i,x})}}{\lVert\mathbf{p}_i-\mathbf{X}_x\rVert^2+\varepsilon^2},
  \label{eq:lamp-direct}
\end{equation}
where $\boldsymbol{\omega}_{i,x}$ is the unit emission direction from the lamp to $\mathbf{X}_x$, and $\varepsilon$ a learned softening that keeps the inverse square finite next to the lamp.
The lamp's 3D position $\mathbf{p}_i$ comes from depth: the mask's centre at its median depth locates the lamp's shade, and the predicted offset corrects it to the actual light (appendix, Sec.~\ref{supp:lamp}).

\paragraph{Lobe.}
A shaded lamp lights the room evenly through its fabric and throws stronger cones out of its openings.
We model this as
\begin{equation}
  f_i(\boldsymbol{\omega}) = w^{\mathrm{iso}}_i + \tfrac{\kappa_i+1}{3}
  \big( w^{\mathrm{dn}}_i \pos{\mathbf{a}\!\cdot\boldsymbol{\omega}}^{\kappa_i}
      + w^{\mathrm{up}}_i \pos{-\mathbf{a}\!\cdot\boldsymbol{\omega}}^{\kappa_i} \big),
  \label{eq:lobe}
\end{equation}
where $\boldsymbol{\omega}$ is any emission direction ($\boldsymbol{\omega}_{i,x}$ in \cref{eq:lamp-direct}) and $\mathbf{a}$ points down in the image.
The weights $w^{\mathrm{iso}}_i$, $w^{\mathrm{dn}}_i$ and $w^{\mathrm{up}}_i$ scale the isotropic term and the downward and upward cones;
the exponent $\kappa_i$, fixed per lamp type, sets how narrow the cones are, and $(\kappa_i+1)/3$ keeps the emitted power independent of it
(appendix, Secs.~\ref{supp:lamp} and~\ref{supp:ablations}).

\subsubsection{Windows}
\label{sec:windows}

\paragraph{Parameters.} A window $j\in\Wset$ lets daylight in, so in the photograph it is on ($o_j=1$), and its
colour and intensity are read from its pixels: its intensity $\Phi_j=\bar Y_j$ is the mean linear pixel value over the window mask, and its colour
$c_j$ the mean of each channel divided by $\bar Y_j$.
Its emission, how the window appears in the image, is $E_{j,x}=M_{j,x}\,\gamma\,\hat\psi_x$, where $\hat\psi$, predicted like $\hat\xi$,
is the pattern of the view through the window and $\gamma$ one learned gain shared by all windows.

\paragraph{Direct light.}
We approximate the skylight entering through a window as a diffuse emitter spanning its opening, with its intensity tied to the recorded brightness of its window mask.
We split the window into small elements $e$, each with a 3D area $w_{j,e}$, centroid $\mathbf{X}_{j,e}$ and normal $\mathbf{n}_{j,e}$ (appendix, Sec.~\ref{supp:winarea}).

The direct light at pixel $x$ is the sum of the point-to-element form factors~\cite{cohen1993radiosity} from $x$ to the centroid of each
element, scaled by a fixed gain $\beta$ (appendix, Sec.~\ref{supp:winarea}),
\begin{align}
  D_{j,x} &= \beta \sum_{e} F_j(x,e),
  \label{eq:window-direct}\\
  F_j(x,e) &= \frac{w_{j,e}}{\pi}\,
              \frac{\pos{\mathbf{n}_x\!\cdot\mathbf{u}}\,\pos{-\mathbf{n}_{j,e}\!\cdot\mathbf{u}}}
                   {\lVert\mathbf{X}_{j,e}-\mathbf{X}_x\rVert^2},
  \label{eq:kernel}
\end{align}
where $\mathbf{X}_x$ and $\mathbf{n}_x$ are the 3D point and normal of pixel $x$, and $\mathbf{u}$ is the unit vector from
$\mathbf{X}_x$ to $\mathbf{X}_{j,e}$.

\subsection{Interreflection}
\label{sec:learned}
\label{sec:interreflection}

A large share of the indoor light reaching a surface has bounced at least once. Computing this bounced light physically would need the whole room's geometry and albedo, including what is out of view, so we approximate it and learn its parameters.
Each light $i$ has its own indirect term $R_i$, the product of a \emph{magnitude}, how much of its light bounces, and a
\emph{spatial pattern}, where it lands.
The magnitude is set by the room's effective mean albedo $\bar\rho$, the albedo of an equivalent integrating
sphere~\cite{moon1940interreflections}: a source whose mean direct light over the room is $\avg{D}$ casts a mean total of
$\avg{D}(1+\bar\rho+\bar\rho^2+\cdots)=\avg{D}/(1-\bar\rho)$, of which the part that has bounced at least once is
\begin{equation}
  \frac{\avg{D}}{1-\bar\rho} - \avg{D} = \eta(\bar\rho)\, \avg{D}, \qquad \eta(\bar\rho)=\frac{\bar\rho}{1-\bar\rho},
  \label{eq:sphere}
\end{equation}
where $\eta$ is the interreflection ratio.
We predict $\bar\rho\in[0.3,0.5]$ per scene from the encoder's room code $\mathbf{z}$ (appendix, Sec.~\ref{supp:interrefl}), so
$\eta\in[0.43,1]$, typical of indoor rooms.
The pattern $P_i$ is a unit-mean field predicted from the image: a uniform term plus a virtual directional light, both
varying across the image (appendix, Sec.~\ref{supp:interrefl}). Together,
\begin{equation}
  R_{i,x} = \eta(\bar\rho)\,\avg{D_i}_i\, P_{i,x},
  \label{eq:indirect}
\end{equation}
where $\avg{\cdot}_i$ averages over pixels outside light $i$'s mask and the window mask.

\subsection{Learning per-light shading without per-light labels}
\label{sec:training}

Our method is supervised only by the total shading $S$ of \cref{eq:cast}; per-light shading is never observed.
Each light's layer is instead constrained by its physical model: it is bright where a surface is near the light and faces
it, and fades with distance.
Fitting the total thus makes each light explain the part of the room it can reach, and a user can edit the fitted parameters.

\paragraph{Supervision without human labels.}
We train on $83$k LSUN bedroom photographs~\cite{yu2015lsun} at $256\times256$.
No photo is human-annotated: masks come from SAM~3~\cite{carion2025sam3}, depth and normals from
Metric3D~\cite{hu2024metric3dv2}, and the shading target $S^{\mathrm{T}}$
is the diffuse shading of a pretrained intrinsic model~\cite{careaga2024colorful}, run on our frames.
No multi-light images, rendered images, synthesised pseudo-pairs~\cite{yen2026freelit,cui2022local} or manual labels enter
our training, though the off-the-shelf models we use were trained on some of them. Each gives only per-image outputs (one total shading, one depth map, one set of masks), none of which separates the lights: the split among lights is learned from single images alone.
The shading target $S^{\mathrm{T}}$ carries albedo detail, but our shading is computed from light models and geometry rather than regressed per pixel,
which keeps most of it out (\cref{fig:leak}).
On/off labels for lamps come from thresholding the photograph's own brightness, with no per-image annotation (appendix, Sec.~\ref{supp:onoff}).

\paragraph{Shading loss.}
The main loss compares log shading rather than shading, so bright regions do not dominate the loss:
\begin{equation}
  \mathcal{L}_{S} = \frac{1}{\lvert\Omega\rvert}\sum_{x\in\Omega} \big\lvert\log S_x - \log S^{\mathrm{T}}_x\big\rvert,
  \label{eq:loss-main}
\end{equation}
where $\Omega$ is the set of pixels outside every window mask; window pixels are
excluded because the target cannot separate the view through a window from the room's shading.

The total loss adds a window term, a focal on/off loss~\cite{lin2017focal} on the labelled lamps and an off-state term that
makes a switched-off lamp look like the surfaces next to it (weights and definitions in the appendix, Sec.~\ref{supp:s1train}).

\paragraph{Network.}
The extrinsic encoder is a ResNet-18~\cite{he2016resnet} with RoI pooling over each lamp's mask ($11.3$\,M parameters); the
emission patterns, the per-scene shading and the interreflection fields come from small CNNs ($1.5$\,M parameters in all;
training details in the appendix, Sec.~\ref{supp:s1train}).

\subsection{Shadows}
\label{sec:visibility}

The light models of \cref{sec:lightmodels} light every surface without considering occlusion.
We therefore cast rays against a mesh lifted from the estimated depth.
Each light's direct light is multiplied by its visibility $V_i$, the share of rays from a pixel to points on the light that
the mesh does not block: replacing $D_i$ by $V_i D_i$ in \cref{eq:formation} gives the shadowed layer $\tilde s_i$, with the bounced light $R_i$ unchanged
(appendix, Sec.~\ref{supp:visibility}), and the rendered shading is $\tilde S=\sum_i \tilde s_i$.
Visibility is computed once per photograph at inference; every later edit reuses it.

\subsection{Editing}
\label{sec:editing}

Our method supports physically based editing of a real photograph's own lights: switching a light on or off, dimming
or recolouring it, one light or several at a time.
An edit changes the parameters of each edited light, and the relit image is the edited shading multiplied by the albedo in linear RGB (appendix, Sec.~\ref{supp:albedo}).
Edits thus change only the light: no object can be hallucinated or removed, as can happen when an image generator
re-synthesises every pixel.

%% file: experiments.tex
\section{Experiments}
\label{sec:experiments}

\subsection{Setup}
\label{sec:setup}

\paragraph{Paired photographs, for testing only.}
We photographed $19$ scenes, by day and at night, each with several sets of its lights switched on, and labelled
every light's state ($59$ photographs, \cref{fig:benchmark}; camera and rooms in the appendix, \cref{supp:geometry}).
They are used only for testing: every ordered pair of photographs of a scene whose labelled states differ is a test pair, $128$
in all, excluding pairs in which every labelled light is off in one photograph, since our model then has no visible light to edit
(\cref{sec:formation}).

\begin{figure*}[t]
  \centering
  \includegraphics[width=0.84\linewidth]{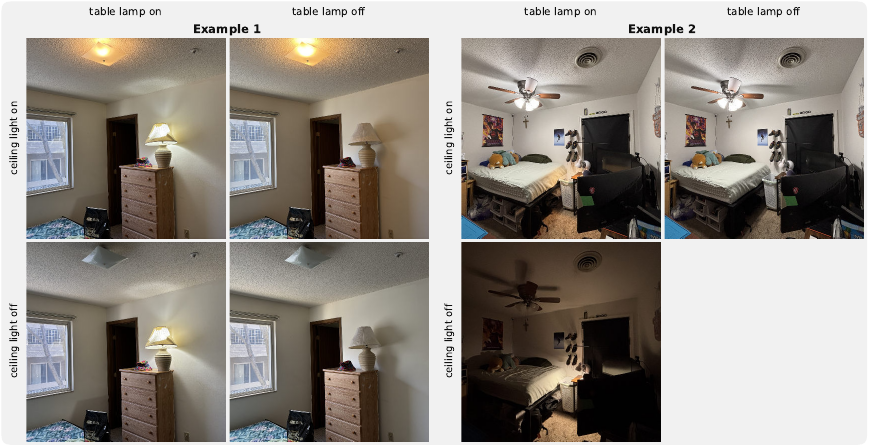}
  \caption{\textbf{Examples of our paired photographs}, each captured under combinations of lights (rows: ceiling light on, off; columns: table lamp on, off); the blank cell, with every light off, is not a test case, since our model then has no visible light to edit (\cref{sec:formation}). Any two photographs of a scene whose lights differ form a test pair.}
  \label{fig:benchmark}
\end{figure*}
\begin{figure*}[t]
  \centering
  \includegraphics[width=\linewidth]{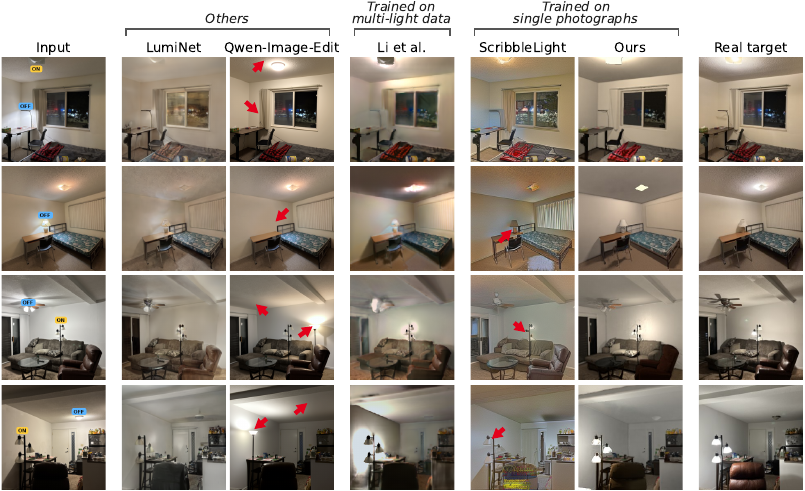}
  \caption{\textbf{Single-image relighting on real photograph pairs.} Four hand-picked pairs, grouped as in \cref{tab:vs-unedited} (Li \etal as published); learned only from LSUN bedrooms, our method carries over to other indoor scenes, such as the kitchen of the last row; eight more, drawn at random, are in \cref{fig:random-pairs}. The lights to be switched are tagged ON or OFF on the input for display only; every method receives the unmarked photograph. Red arrows mark the fixtures an edit hallucinates, removes or replaces.}
  \label{fig:qual}
\end{figure*}

\begin{figure*}[t]
  \centering
  \includegraphics[width=\linewidth]{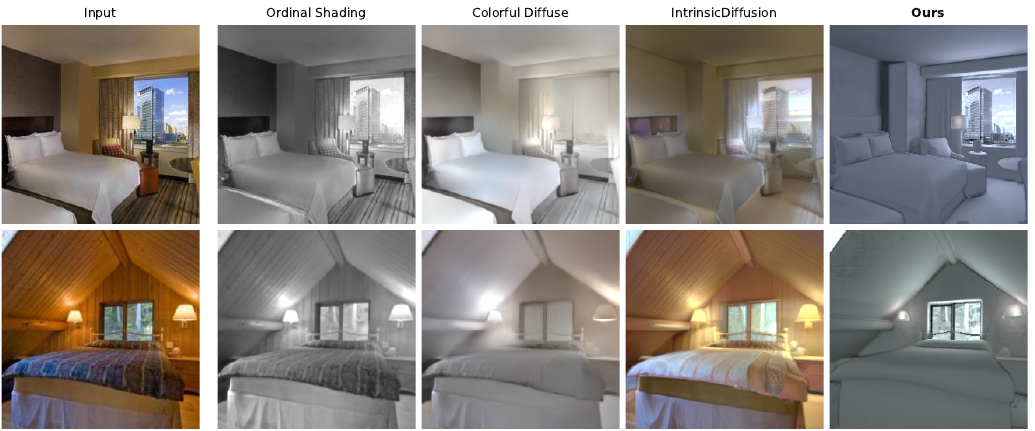}
  \caption{\textbf{Albedo leakage in shading on real photographs.} A shading should depend only on the light and the geometry; surface patterns and colours in it are reflectance that has leaked in.
  The intrinsic shadings~\cite{careaga2023ordinal,careaga2024colorful,luo2024intrinsicdiffusion}, including that of Colorful Diffuse, the model we learn from, carry the carpet's stripes (top) or the wood grain and the quilt's pattern (bottom); ours largely does not.}
  \label{fig:leak}
\end{figure*}
\begin{figure*}[t]
  \centering
  \includegraphics[width=\linewidth]{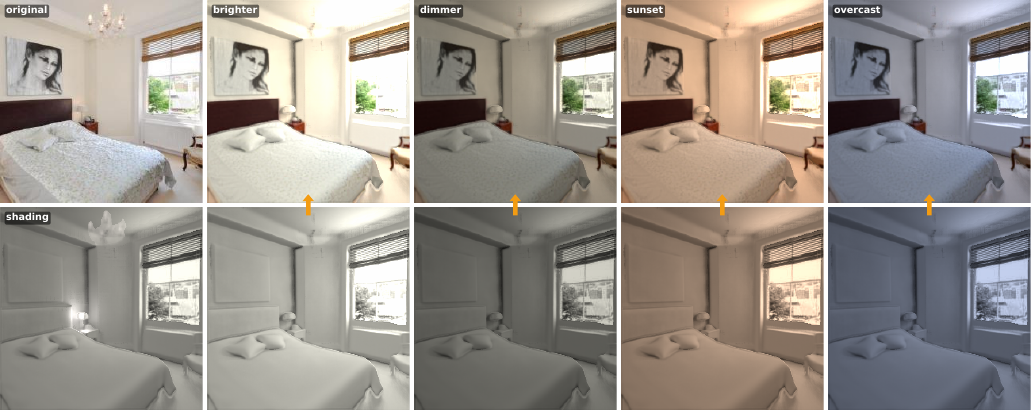}
  \caption{\textbf{Window light editing.} Daylight is a light like any other (with the lamps switched off: the window brightened, dimmed and turned to sunset and overcast light). Each edit is rendered from the albedo and the edited layers, above its shading.}
  \label{fig:window-edit}
\end{figure*}
\begin{figure*}[p]
  \centering
  \includegraphics[width=\linewidth]{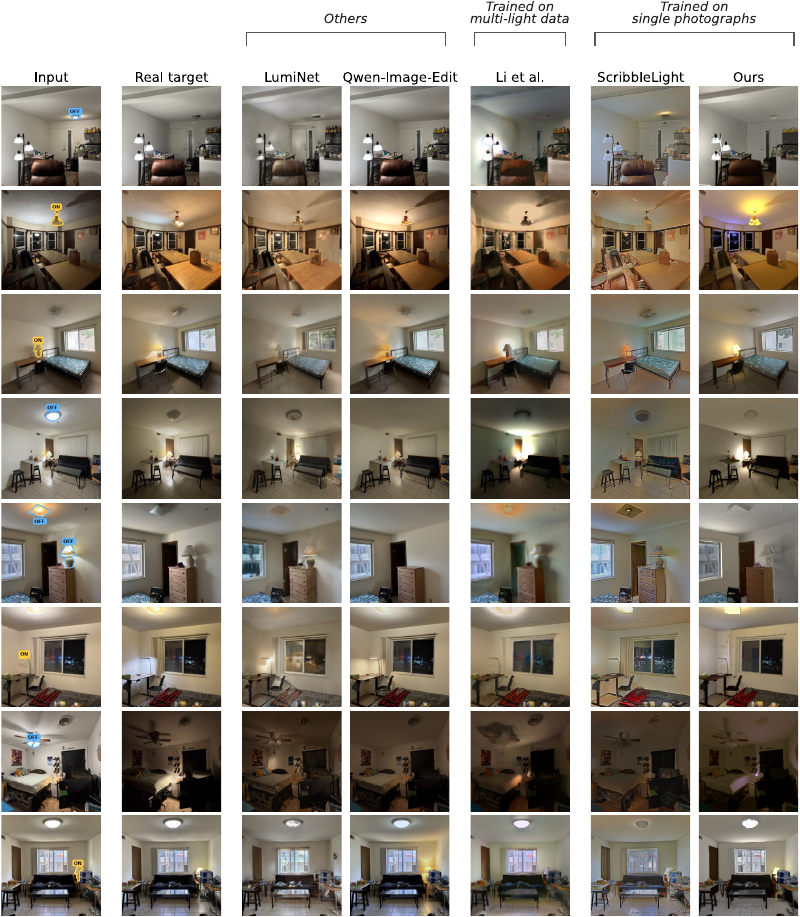}
  \caption{\textbf{Eight random pairs.} Drawn at random from the $128$ test pairs (excluding those of \cref{fig:qual}; rows 1 and 6 are a second random draw, and row 7 is drawn at random from a room not otherwise shown), shown as in \cref{fig:qual} with the real target next to the input: grouped as in \cref{tab:vs-unedited}, ours tuned, Li \etal as published. The input marks the lights to switch, outlined from the masks of \cref{supp:masks}; the outlines and the ON / OFF tags are for display only, and every method receives the unmarked photograph.}
  \label{fig:random-pairs}
\end{figure*}
\paragraph{Protocol.}
Every method receives the same $256\times256$ photograph.
Since the brightness of a lamp switched on by the edit is unknown a priori, we follow LuxRemix~\cite{liang2026luxremix} and fit
light parameters to the target: for methods that support per-light editing, one RGB gain per light, with
switched-off lights kept off (\emph{tuned}, $^\ddagger$; untuned results in the appendix, Sec.~\ref{supp:tuning}).
Every output, including the unedited input, is then scaled per colour channel to the target by least squares, which removes exposure differences between photographs~\cite{farinha2026pixlrelight,liang2026luxremix},
and PSNR, SSIM and LPIPS~\cite{zhang2018unreasonable} are computed on the whole sRGB image and averaged per scene, then over
the $19$ scenes.

\paragraph{Baselines.}
Li \etal~\cite{li2022physically}, the other physically based method, receives the light masks (\cref{fig:qual} shows it as published).
ScribbleLight~\cite{scribblelight} receives scribbles that paint each light white or black by its target state.
LumiNet~\cite{xing2025luminet} receives the target photograph as its lighting reference, which no real edit would have.
Qwen-Image-Edit~\cite{wu2025qwenimage}, a general-purpose image editor, receives a text prompt naming each light to switch.
Ours and ScribbleLight are the only methods compared that never see a room under more than one lighting in training, and ours
and Li \etal the only ones that model each light explicitly (\cref{tab:intro-methods}).

\subsection{Comparisons}
\label{sec:comparisons}
\input{tab_vs_unedited}

\paragraph{By image metrics, most edits make the photograph worse.}
People, in contrast, prefer our edits to the unedited photograph, which ignores the requested change (user study below).
Under LPIPS, which follows human judgements of image similarity~\cite{zhang2018unreasonable}, Li \etal and ScribbleLight come closer to the target than the unedited photograph on $0\%$ of pairs and ours on $12\%$, rising to $47\%$ ($15$ of $32$) on the quarter of pairs where the switch changes the room most (appendix, \cref{fig:scatter-lpips}); ours also has the lowest mean LPIPS of the methods trained on single photographs or on multi-light data ($0.255$, \cref{tab:vs-unedited}).
Like Li \etal, ours re-renders the image from an albedo and per-light shadings, so an edit changes only the light, but fine detail is lost in re-rendering, which the metrics penalise heavily: on the quarter of each image that the switch changes most, ours is $3.2$\,dB closer to the target than the unedited photograph and closer on $87\%$ of pairs, while on the rest, re-rendering costs about $5$\,dB (appendix, \cref{tab:supp-region}).
Qwen-Image-Edit avoids the penalty by largely reproducing the photograph's pixels, with an LPIPS that stays close to the unedited photograph's (appendix, \cref{fig:gain-curves}), which is why it comes closer on $46\%$ of pairs.
But neither it nor ScribbleLight models the lights physically, and they hallucinate, remove or replace a light fixture in $73\%$ and $30\%$ of pairs, respectively (\cref{tab:vs-unedited}; \cref{fig:qual}; appendix, \cref{fig:qwen}).
We list Qwen-Image-Edit, which alters fixtures in most pairs, under \emph{Others} (\cref{tab:vs-unedited,tab:lux-lightlab}).

\paragraph{Comparison with ScribbleLight and Li \etal}
Against ScribbleLight, trained like ours on single photographs, our edits have the lower LPIPS on $95\%$ of pairs.
Against Li \etal, physically based like ours but trained on multi-light data, ours has the lower LPIPS on $77\%$ of pairs and the lower mean LPIPS in $17$ of the $19$ scenes.

\paragraph{Comparison with LuxRemix and LightLab.}
On their own data, LuxRemix~\cite{liang2026luxremix} and LightLab~\cite{lightlab}, trained on multi-light data, outperform ours (\cref{tab:lux-lightlab}; qualitative comparison in the appendix, \cref{fig:supp-lux-lightlab}).
Neither releases its model, so neither can be tested on the $128$ test pairs.
The only real photographs all methods share are the $4$ eligible pairs among the examples LightLab released, whose outputs LuxRemix's authors shared (appendix, Secs.~\ref{supp:luxremix} and~\ref{supp:lightlab}).
Both sets favour them.
LuxRemix's test rooms come from the same generator as its training renders, whereas ours learns only from real photographs (\cref{tab:lux-lightlab}).
On LightLab's $4$ real pairs, no method improves on the unedited photograph under SSIM or LPIPS, LuxRemix and LightLab included (\cref{tab:lux-lightlab}).
Multi-light training data helps but is not always available; among methods trained without it, ours is ahead of ScribbleLight on every metric on both datasets (\cref{tab:lux-lightlab}).
LuxRemix's cases also test each light's layer directly, since each shows the room lit by one light alone: there, ours is closer to the target than Li~\etal on $85$ of the $112$ lights, although only Li~\etal is trained with per-light ground truth (appendix, Sec.~\ref{supp:luxremix}).

\input{tab_lux_lightlab}
\input{tab_ablation}
\paragraph{User study.}
People prefer our edits to other methods' edits and to the unedited photograph.
Image metrics score an edit by its similarity to one target photograph, which favours leaving the photograph unedited: switching a light changes only part of the image, and the metrics also penalise fine detail lost in re-rendering.
In the end, people judge whether an edit is good, so we asked them (details in the appendix, Sec.~\ref{supp:userstudy}).
Against other methods, asked which image looks more like a real photograph of the room after the change, participants chose our edit over Li~\etal's in $84\%$ of comparisons and over ScribbleLight's in $93\%$ (\cref{tab:vs-unedited}).
Against the unedited photograph, asked which image best represents the room after the change, $13$ participants chose our edit in $92\%$ of $151$ comparisons: people reward the change they asked for, even an imperfect one.

\paragraph{Editing.}
Beyond the switches a pair can test, our method supports any edit of a light's parameters (\cref{fig:teaser,fig:window-edit}): dimming, recolouring and chained edits, for lamps and windows. More photos for each control (\cref{fig:supp-controls}), more scenes' layers (\cref{fig:decomp-gallery}) and direct sunlight cast through a window at inference (\cref{supp:controls}) are in the appendix and the supplementary video.

\paragraph{Ablations.}
We evaluate our model with one part removed at a time (\cref{tab:ablation}): interreflection is essential, and ray-traced shadows help modestly.
Interreflection carries much of a room's light: retrained without it, PSNR falls by $2.9$\,dB and LPIPS rises from $0.255$ to $0.368$.
The ray-traced shadows add only $0.4$\,dB and leave SSIM and LPIPS unchanged: the shadows they cast cover little of the image ($V_i$ in \cref{fig:overview}), so whole-image metrics barely register them, even where a viewer would notice.
Letting each light's interreflection ignore the edit, like an ambient term, costs $0.6$\,dB of PSNR (\cref{tab:ablation}); this row, the model without both parts and what the shadows change are discussed in the appendix (Sec.~\ref{supp:ambient}).

\paragraph{Albedo leakage.}
\label{sec:leak-cost}
Our shading carries far less of the colours and patterns of textured surfaces than recent intrinsic decompositions, including the one it learns from (\cref{fig:leak}): it is computed from per-light physical models and ray casting rather than regressed from pixels.
On SAW~\cite{kovacs2017saw}, it scores a challenge AP~\cite{li2018cgintrinsics} of $94.64$, against $96.48$ for the model it learns from (appendix, \cref{tab:saw}).
The gap lies at shadow boundaries ($68.2$ against $90.4$): our shadows are ray-cast against the estimated geometry, which is less exact than copying them from the photograph, but each light's shadows are switched and dimmed with it.

\paragraph{Size and speed.}
Our network has $12.8$\,M parameters, $35\times$ fewer than the intrinsic model it learns from.
A photograph is decomposed once; each edit then only re-weights physically modelled layers and takes about $0.1$\,ms, against $4$--$5$\,s for the diffusion relighters, which run again for every edit (\cref{tab:intro-methods}; appendix, \cref{fig:cost}).

%% file: tab_vs_unedited.tex
\providecommand{\hcell}[1]{\begin{tabular}[b]{@{}c@{}}#1\end{tabular}}
\begin{table}[!t]
\caption{\textbf{Single-image relighting.} Our method keeps every fixture intact (unlike Qwen-Image-Edit and ScribbleLight), is trained on single images (unlike Li \etal), has the best SSIM and LPIPS of the methods trained on single photographs or on multi-light data and wins most human comparisons, on the $128$ test pairs (\cref{sec:setup}). $^\ddagger$Tuned (untuned results in the appendix, \cref{tab:supp-tuning}); $^\dagger$sees the target; bold: best per training group. \emph{Fixtures intact}: pairs in which no light fixture is hallucinated, removed or replaced (close-ups in the appendix, \cref{fig:qwen}). \emph{Our win rate}: user-study preference rates for ours against that row, judged on $^a$realism or $^b$fidelity to the change (\cref{sec:comparisons}).}
\label{tab:vs-unedited}
\centering
\footnotesize
\setlength{\tabcolsep}{3pt}
\begin{tabular}{@{}lccccc@{}}
\toprule
Method                                 & \hcell{Fixtures\\intact$\uparrow$} & PSNR$\uparrow$ & SSIM$\uparrow$ & LPIPS$\downarrow$ & \hcell{Our win\\rate$\uparrow$} \\
\midrule
\multicolumn{6}{@{}l}{\emph{Trained on single photographs}} \\
Ours$^\ddagger$                        & \textbf{100\%} & \textbf{18.66} & \textbf{0.734} & \textbf{0.255} & -- \\
ScribbleLight~\cite{scribblelight}     & $70\%$ & 15.93 & 0.569 & 0.349 & $93\%^a$ \\
\midrule
\multicolumn{6}{@{}l}{\emph{Trained on multi-light data}} \\
Li \emph{et al.}$^\ddagger$~\cite{li2022physically} & \textbf{100\%} & \textbf{18.99} & \textbf{0.731} & \textbf{0.308} & $84\%^a$ \\
\midrule
\multicolumn{6}{@{}l}{\emph{Others}} \\
\textit{LumiNet}$^\dagger$~\cite{xing2025luminet} & \textit{100\%} & \textit{17.52} & \textit{0.748} & \textit{0.235} & -- \\
\textit{Qwen-Image-Edit}~\cite{wu2025qwenimage} & \textit{27\%} & \textit{20.86} & \textit{0.833} & \textit{0.161} & -- \\
\textit{Unedited photo}                & -- & \textit{18.57} & \textit{0.813} & \textit{0.159} & $92\%^b$ \\
\bottomrule
\end{tabular}

\end{table}

%% file: tab_lux_lightlab.tex
\providecommand{\tbd}{\ifdefined\textcolor\textcolor{red}{?}\else ?\fi}
\providecommand{\hcell}[1]{\begin{tabular}[b]{@{}c@{}}#1\end{tabular}}
\begin{table}[t]
\caption{\textbf{On LuxRemix's and LightLab's data.} The methods trained on multi-light data lead on their own data; among methods trained on single photographs, ours is best on every metric on both (appendix, Secs.~\ref{supp:luxremix} and~\ref{supp:lightlab}; examples in \cref{fig:supp-lux-lightlab}). $^*$As reported, under its own scoring. Marks and groups as in \cref{tab:vs-unedited}; underlined: second among multi-light methods.}
\centering
\footnotesize
\setlength{\tabcolsep}{2.2pt}
\providecommand{\luxbox}{}\newsavebox{\luxtabbox}
\sbox{\luxtabbox}{\begin{tabular}{@{}lcccccc@{}}
\toprule
 & \multicolumn{3}{c}{\hcell{LuxRemix: per light\\($112$ renders)}} & \multicolumn{3}{c}{\hcell{LightLab: relighting\\($4$ real pairs)}} \\
\cmidrule(lr){2-4}\cmidrule(l){5-7}
Method & PSNR$\uparrow$ & SSIM$\uparrow$ & LPIPS$\downarrow$ & PSNR$\uparrow$ & SSIM$\uparrow$ & LPIPS$\downarrow$ \\
\midrule
\multicolumn{7}{@{}l}{\emph{Trained on single photographs}} \\
Ours                                   & \textbf{17.46} & \textbf{0.691} & \textbf{0.328} & \textbf{15.84}$^\ddagger$ & \textbf{0.613}$^\ddagger$ & \textbf{0.500}$^\ddagger$ \\
ScribbleLight~\cite{scribblelight}     & 16.76 & 0.570 & 0.451 & 14.52 & 0.368 & 0.535 \\
\midrule
\multicolumn{7}{@{}l}{\emph{Trained on multi-light data}} \\
Li \etal~\cite{li2022physically}       & \underline{16.85} & \underline{0.658} & \underline{0.382} & 17.08$^\ddagger$ & 0.660$^\ddagger$ & 0.472$^\ddagger$ \\
LightLab~\cite{lightlab}               & --    & --    & --    & \textbf{19.25} & \underline{0.754} & \textbf{0.147} \\
LuxRemix~\cite{liang2026luxremix} & \textbf{27.68}$^*$ & \textbf{0.898}$^*$ & \textbf{0.082}$^*$ & \underline{18.75} & \textbf{0.781} & \underline{0.150} \\
\midrule
\multicolumn{7}{@{}l}{\emph{Others}} \\
\textit{LumiNet}$^\dagger$~\cite{xing2025luminet}          & \textit{17.59} & \textit{0.727} & \textit{0.262} & \textit{18.19} & \textit{0.769} & \textit{0.198} \\
\textit{Qwen-Image-Edit}~\cite{wu2025qwenimage}       & \textit{17.32} & \textit{0.742} & \textit{0.247} & \textit{20.12} & \textit{0.755} & \textit{0.167} \\
\textit{Unedited input}                & \textit{18.49} & \textit{0.769} & \textit{0.206} & \textit{18.46} & \textit{0.788} & \textit{0.144} \\
\bottomrule
\end{tabular}}
\ifdim\wd\luxtabbox>\linewidth\resizebox{\linewidth}{!}{\usebox{\luxtabbox}}\else\usebox{\luxtabbox}\fi
\label{tab:lux-lightlab}
\end{table}

%% file: tab_ablation.tex
\providecommand{\tbd}{\ifdefined\textcolor\textcolor{red}{?}\else ?\fi}
\begin{table}[t]
\caption{\textbf{Ablations.} On the $128$ test pairs, tuned and scored as in \cref{tab:vs-unedited}. \emph{w/o interreflection} is retrained without it; the other rows change the model at inference.}
\label{tab:ablation}
\centering
\small
\setlength{\tabcolsep}{5pt}
\begin{tabular}{@{}lccc@{}}
\toprule
Variant & PSNR$\uparrow$ & SSIM$\uparrow$ & LPIPS$\downarrow$ \\
\midrule
Ours                                    & \textbf{18.66}    & \textbf{0.734} & \underline{0.255} \\
w/o ray-traced shadows                  & \underline{18.28} & \textbf{0.734} & \underline{0.255} \\
w/o interreflection                     & 15.71             & 0.597          & 0.368 \\
w/o interreflection and shadows         & 16.04             & 0.651          & 0.320 \\
Interreflection ignores the edit        & 18.05             & \underline{0.713} & \textbf{0.248} \\
\bottomrule
\end{tabular}
\end{table}

%% file: conclusion.tex
\section{Limitations}
\label{sec:limitations}
While explicit per-light models make every edit controllable, our approach has remaining limitations.
First, an edit re-renders the photograph as albedo times shading, so fine detail is lost even where the light does not change (\cref{sec:comparisons}), and the model runs at $256\times256$, the resolution of its training photographs.
Second, each window is a light of its own, dimmed, recoloured or switched off separately rather than as part of an ambient term, but we model its light as diffuse skylight only; direct sunlight can be cast through a window at inference, in a chosen direction (appendix, \cref{supp:controls}).
Third, where a bright window lights the same surfaces as the lamps, it can take some of their light.
Failure cases are in the appendix (\cref{fig:supp-failures}).

\section{Conclusion}
\label{sec:conclusion}
We introduced PLRS, a per-light reshading method that learns from single images.
We represent a photograph's luminaires themselves rather than their effect on the image: we recover physical light models of the photograph's lamps, ceiling fixtures and windows and relight it by editing their parameters.
To our knowledge, PLRS is the first such method.
Each light has a physically modelled shading layer, so an edit switches, dims or recolours one light in about $0.1$\,ms and never hallucinates, removes or replaces a fixture.
People prefer our edits to those of Li \etal and ScribbleLight, and to leaving the photograph unedited.
We believe that physical light models learned from single images are a practical route to light editing wherever multi-light data cannot be had; recovering fine detail without giving up per-light control is a natural next step.

%% file: X_suppl.tex
\clearpage
\twocolumn[\centering\Large\textbf{Appendix}\\\vspace{1.0em}]
\appendix
\renewcommand{\thesection}{S}
\renewcommand{\thesubsection}{S\arabic{subsection}}
\makeatletter\renewcommand*{\Hy@backout}[1]{}\makeatother
\input{supplementary}

%% file: supplementary.tex
\setcounter{topnumber}{3}\setcounter{dbltopnumber}{3}\renewcommand{\topfraction}{0.9}\renewcommand{\dbltopfraction}{0.9}%
\renewcommand{\textfraction}{0.07}\renewcommand{\floatpagefraction}{0.6}\renewcommand{\dblfloatpagefraction}{0.6}

This appendix gives method details, training details and design choices, evaluation details and further comparisons, and more results and failure cases, in that order; the supplementary video shows edits being made (\cref{supp:controls}).
Each section points to the part of the paper it details.
\noindent\textbf{Contents}\par
{\small
\noindent\emph{Method details}\par
\noindent\hspace*{1em}\ref{supp:geometry}\enspace Camera and geometry\dotfill\pageref{supp:geometry}\par
\noindent\hspace*{1em}\ref{supp:masks}\enspace Light-source masks\dotfill\pageref{supp:masks}\par
\noindent\hspace*{1em}\ref{supp:lamp}\enspace Lamp parameters, position and lobe\dotfill\pageref{supp:lamp}\par
\noindent\hspace*{1em}\ref{supp:winarea}\enspace Window elements and area per pixel\dotfill\pageref{supp:winarea}\par
\noindent\hspace*{1em}\ref{supp:interrefl}\enspace Interreflection\dotfill\pageref{supp:interrefl}\par
\noindent\hspace*{1em}\ref{supp:fallback}\enspace Rooms without an active light\dotfill\pageref{supp:fallback}\par
\noindent\hspace*{1em}\ref{supp:visibility}\enspace Ray-traced visibility\dotfill\pageref{supp:visibility}\par
\noindent\hspace*{1em}\ref{supp:albedo}\enspace The albedo\dotfill\pageref{supp:albedo}\par
\noindent\emph{Training details and design choices}\par
\noindent\hspace*{1em}\ref{supp:onoff}\enspace On/off pseudo-labels\dotfill\pageref{supp:onoff}\par
\noindent\hspace*{1em}\ref{supp:s1train}\enspace Losses and optimisation\dotfill\pageref{supp:s1train}\par
\noindent\hspace*{1em}\ref{supp:ambient}\enspace Ablations of shadows and interreflection\dotfill\pageref{supp:ambient}\par
\noindent\hspace*{1em}\ref{supp:ablations}\enspace Fixed parameters\dotfill\pageref{supp:ablations}\par
\noindent\emph{Evaluation on the paired photographs}\par
\noindent\hspace*{1em}\ref{supp:tuning}\enspace Tuned and untuned\dotfill\pageref{supp:tuning}\par
\noindent\hspace*{1em}\ref{supp:gain}\enspace More comparisons on the paired photographs\dotfill\pageref{supp:gain}\par
\noindent\hspace*{1em}\ref{supp:userstudy}\enspace User study\dotfill\pageref{supp:userstudy}\par
\noindent\emph{Evaluation on LuxRemix's and LightLab's data}\par
\noindent\hspace*{1em}\ref{supp:luxremix}\enspace Evaluation on the LuxRemix test set\dotfill\pageref{supp:luxremix}\par
\noindent\hspace*{1em}\ref{supp:lightlab}\enspace The LightLab pairs\dotfill\pageref{supp:lightlab}\par
\noindent\emph{More results}\par
\noindent\hspace*{1em}\ref{supp:controls}\enspace Editing each light\dotfill\pageref{supp:controls}\par
\noindent\hspace*{1em}\ref{supp:leakcost}\enspace Shading quality, size and speed\dotfill\pageref{supp:leakcost}\par
\noindent\hspace*{1em}\ref{supp:failures}\enspace Failure cases\dotfill\pageref{supp:failures}\par
}

\section*{Method details}

\subsection{Camera and geometry}
\label{supp:geometry}

We first recover the photograph's 3D geometry (\cref{sec:scene}): each pixel $x=(u,v)$ is back-projected with a pinhole camera, $\mathbf{X}_x=\tilde d_x K^{-1}[u,v,1]^{\top}$, where $\tilde d$ is the estimated depth divided by its median.
$K$ has square pixels and its principal point at the image centre, so the vertical field of view $\theta$ alone sets it: the focal length is $f=(H/2)/\tan(\theta/2)$ pixels for an image of height $H$.
LSUN provides no camera intrinsics, so we assume $\theta=55^\circ$, typical of interior photography (about a
23\,mm lens on a full-frame camera); the photographs are centre-cropped, not resized, to a square.
The paired photographs, all taken with one phone (an iPhone~15) on a tripod, use the lens's own $\theta$: $90^\circ$ at $0.5\times$ and $53^\circ$ at $1\times$.
The scene's metric scale is not needed: the median normalisation removes it, together with the focal length that converts Metric3D's
canonical-camera depth to metres, and any remaining per-image scale is absorbed by the light intensities.
Each of the $19$ scenes of the paired photographs is a room photographed from one camera angle. \Cref{tab:supp-dataset} summarises them.

\subsection{Light-source masks}
\label{supp:masks}

Every light needs a mask and a class (\cref{sec:scene}). We run SAM~3~\cite{carion2025sam3} once per text prompt, keep the instances with a detection score of at least 0.5
(mask threshold 0.5), and unite each prompt's instances into its class:
\begin{itemize}
  \item \textbf{lamp:} ``lamp'', ``floor lamp'', ``table lamp'', ``desk lamp'', ``pendant light'', ``wall sconce'', ``lantern'';
  \item \textbf{ceiling light:} ``ceiling light'', ``chandelier'', ``recessed light'', ``track light'', ``spotlight'', ``light fixture'';
  \item \textbf{window:} ``window'', then more prompts where it finds nothing (below).
\end{itemize}
Of the two light types, lamp and window, a ceiling light is a lamp: it uses the lamp's physical model (\cref{sec:lamps}), and only its fixed cone exponent (\cref{supp:lamp}) and on/off pseudo-label (\cref{supp:onoff}) differ.
\paragraph{False negatives in window masks.}
We noticed that the prompt ``window'' alone misses many daylight openings, glass doors among them.
To recover them, on every photograph where it finds nothing (26,086 photographs), we run SAM~3 again with that prompt and six more: ``window with
curtains'', ``glass door'', ``sliding door'', ``balcony door'', ``french door'' and ``skylight''.
A photograph receives a window mask if the union of these detections covers at least 1\% of the image, unless
``window with curtains'' fired and the plain prompt ``window'' did not: a window covered by curtains, with no glass visible, is not treated as a light source.
The mask is the union of every prompt except ``window with curtains''.
This adds a window mask to 9,110 photographs (8,188 training, 922 validation; example in \cref{fig:supp-masks}).

\paragraph{False positives in ceiling-light masks.}
The ceiling-light prompts return many of them: lamps standing on furniture.
To move these to the lamp class, we test which way the surface around each detection faces.
The room's up axis is the direction most surface normals are parallel to, as floors, ceilings and table tops share it: their first principal axis, taken upward. For every connected component of the ceiling-light mask (at least 20 pixels), we take the median of $\mathbf{n}\cdot\mathbf{up}$ over a ring 1 to 6 pixels around it.
If the ring faces down ($<-0.15$, a ceiling) the component stays a ceiling fixture; if it faces up ($>0.15$, a floor or a
table top) it is moved to the lamps; otherwise it stays a ceiling fixture only if its median height, normalised between the
2nd and 98th percentiles of the room's heights, is at least 0.75.
{\renewcommand{\thefigure}{A}\renewcommand{\theHfigure}{suppA}%
\begin{figure}[t]
  \centering
  \includegraphics[width=\linewidth]{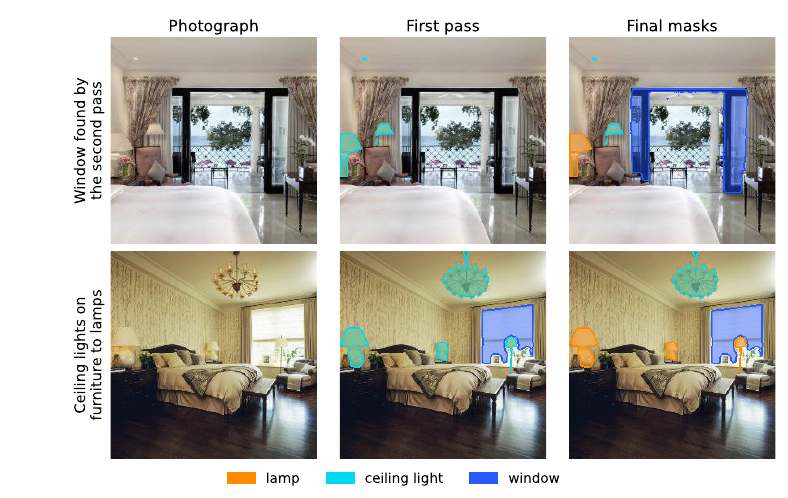}
  \caption{\textbf{Light-source masks} (\cref{supp:masks}) on two LSUN photographs: lamps orange, ceiling lights cyan, windows blue.
  Top: the prompt ``window'' finds nothing, and the second pass finds the glass doors. Bottom: the ceiling-light prompts also fire on the bedside lamps;
  the test on the surface around each detection moves them to the lamps, and the chandelier stays a ceiling light.}
  \label{fig:supp-masks}
\end{figure}}%
\addtocounter{figure}{-1}%
\paragraph{Test photographs.}
Training uses these automatic masks as they are. For the paired photographs and the other test data, the masks are reviewed by hand, with the rules listed in \cref{supp:luxremix}.

\subsection{Lamp parameters, position and lobe}
\label{supp:lamp}

We predict each lamp's parameters (\cref{sec:lamps}) from its per-light feature $\mathbf{h}_i$ with small heads $\ell_{\ast}$:
\begin{equation}
  \begin{aligned}
  o_i &= \mathbf{1}\!\left[\sigma(\ell_o(\mathbf{h}_i))>0.5\right], &
  c_i &= \operatorname{softplus}(\ell_c(\mathbf{h}_i)),\\
  \Phi_i &= \operatorname{softplus}(\ell_\Phi(\mathbf{h}_i)), &
  \gamma_i &= \operatorname{softplus}(\ell_\gamma(\mathbf{h}_i)),
  \end{aligned}
  \label{eq:lamp-params}
\end{equation}
where $o_i$ is the lamp's on/off state, $c_i$ its colour, $\Phi_i$ the learned intensity of its shading and $\gamma_i$ its emission gain.

\paragraph{Position.}
The centre of a lamp's mask need not be precisely where its light comes from, so we learn an offset from that centre, bounded by the lamp's size:
\begin{equation}
  \mathbf{p}_i=\bar{\mathbf{p}}_i+r_i\tanh(\boldsymbol{\delta}_i).
  \label{eq:lamp-position}
\end{equation}
The anchor $\bar{\mathbf{p}}_i=\tilde d_i\,K^{-1}[\bar u_i,\bar v_i,1]^{\top}$ is fixed by the mask and the geometry: $(\bar u_i,\bar v_i)$ is the centroid pixel of the lamp's mask $M_i$,
$K$ the camera of Sec.~\ref{supp:geometry}, and $\tilde d_i$ the median depth over the mask (the scene's median depth for masks
under 131 pixels, whose median is unreliable), so $\bar{\mathbf{p}}_i$ is the 3D point at the centre of the lamp.
The offset is the second term: $r_i$ is the lamp's radius, measured by placing every pixel of
$M_i$ at depth $\tilde d_i$ (its viewing ray scaled to that depth) and taking the median distance of these points to
$\bar{\mathbf{p}}_i$, and $\boldsymbol{\delta}_i\in\mathbb{R}^3$ is predicted
from the lamp feature $\mathbf{h}_i$, so the emitter moves at most $r_i$ along each axis and stays near the lamp.

\paragraph{Normalisation.}
A lamp's brightness should be carried by its intensity parameters alone, so that nothing else competes with them. The emission pattern $\hat\xi$ is therefore normalised to unit mean over the lamp's mask, $\avg{\hat\xi}_{M_i}=1$, so the pattern cannot change how bright the lamp is; a window's pattern $\hat\psi$ is normalised likewise, $\avg{\hat\psi}_{M_j}=1$.
A lamp's colour $c_i$ is normalised to unit mean over the channels, so it carries no brightness.
\paragraph{Lobe.}
The lobe sets where a lamp's light goes. In \cref{eq:lobe}, $\mathbf{a}$ points down in the image, and the three weights are predicted per lamp around a prior of
$(0.31, 2.0, 0.08)$.
The cone width should change where a lamp's light goes, not how much light it emits. A cone $\pos{\mathbf{a}\cdot\boldsymbol{\omega}}^{\kappa}$, with $\pos{t}=\max(t,0)$, emits a total power of
$2\pi\int_0^{\pi/2}\cos^{\kappa}\theta\,\sin\theta\,d\theta=2\pi/(\kappa+1)$ over all directions, so the factor \mbox{$(\kappa+1)/3$} gives every cone the total emitted power of the $\kappa=2$ cone, $2\pi/3$.
We fix $\kappa_i=2$ for ceiling fixtures and $\kappa_i=8$ for other lamps (\cref{supp:ablations}).

\subsection{Window elements and area per pixel}
\label{supp:winarea}

A window is an area light (\cref{sec:windows}): its direct light depends on the window's 3D area and on where that area sits, and this section builds both from the mask. We represent each window as a grid of elements, the standard treatment of an area source. Because masks, depth and
normals are all defined per pixel, we grid the image rather than the window: a $48\times48$ grid of image cells uses these
maps directly, whereas a grid laid on the window would require interpolating them.

Element $e$ of a window consists of the window's pixels inside one cell of the $48\times48$ grid. Its area, centroid and normal are weighted by $\mathcal{A}_y$, the 3D area of the surface seen by pixel $y$, whose normal is $\mathbf{n}_y$, summing over the pixels $y$ of the element:
\begin{equation}
  \begin{aligned}
  w_{j,e} &= \sum_{y\in e} \mathcal{A}_y,\\
  \mathbf{X}_{j,e} &= \frac{1}{w_{j,e}}\sum_{y\in e} \mathcal{A}_y\,\mathbf{X}_y,\\
  \mathbf{n}_{j,e} &= \frac{\sum_{y\in e} \mathcal{A}_y\,\mathbf{n}_y}{\big\lVert\sum_{y\in e} \mathcal{A}_y\,\mathbf{n}_y\big\rVert}.
  \end{aligned}
  \label{eq:window-patch}
\end{equation}
In the form factors, the distance is clamped to at least $r_{\min}$, half the spacing between neighbouring elements, to avoid singularities,
and by convention the form factor between a pixel and
the element in its own cell is zero: each cell approximates its piece of the image with a small plane, and the pixel lies on
that plane.

\paragraph{3D window area.}
Each pixel should count for the glass area it sees. Counting a window's pixels measures how large it looks, which depends on its distance and on the angle we see it at, while
the light it lets in depends on its real glass area.
Weighting each pixel by $\mathcal{A}_y$ makes the sum the real area: of two identical windows, one twice as far away covers a quarter
of the pixels, but each of its pixels sees four times the area, and a window seen nearly edge-on covers few pixels that each
see a long strip of glass.
The 3D area seen by pixel $y$ is
\begin{equation}
  \mathcal{A}_y=\frac{d\omega_y\,\lVert\mathbf{X}_y\rVert^2}{\lvert\cos\theta_y\rvert},
  \qquad
  d\omega_y=\frac{\cos^3\!\alpha_y}{f^2},
  \label{eq:pixel-area}
\end{equation}
where $d\omega_y$ is the solid angle the pixel sees, $\lVert\mathbf{X}_y\rVert$ its distance to the surface it hits and $\theta_y$ the angle between that surface's normal and the pixel's ray;
$f$ is the focal length in pixels and $\alpha_y$ the angle between the ray and the optical axis ($\cos\alpha_y=z_y/\lVert\mathbf{X}_y\rVert$, with $z_y$ the depth).

Both parts come from applying one rule of radiometry twice. A small flat patch of area $dA$, at distance $r$ and tilted by $\theta$ from the line of sight, covers the solid angle~\cite[Sec.~4.2.3, Eq.~4.9]{pharr2023pbrt}
\[
  d\omega=\frac{dA\,\cos\theta}{r^2}.
\]
\emph{Solid angle of a pixel.}
The pixel is such a patch, on the image plane.
Its area is $dA=1$, since lengths on the image plane are measured in pixels.
The image plane lies at distance $f$ from the camera centre along the optical axis, so a pixel whose ray makes the angle $\alpha_y$ with that axis lies at distance $r=f/\cos\alpha_y$.
The image plane faces along the optical axis, so the pixel is tilted by $\theta=\alpha_y$ from its own ray.
The rule gives
\[
  d\omega_y=\frac{\cos\alpha_y}{(f/\cos\alpha_y)^2}=\frac{\cos^3\!\alpha_y}{f^2}.
\]
\emph{Surface inside that solid angle.}
The same cone of directions continues to the scene, and the patch of surface inside it is what the pixel sees.
Its area $\mathcal{A}_y$ is the unknown; its distance is $r=\lVert\mathbf{X}_y\rVert$ and its tilt is $\theta_y$.
The rule now reads
\[
  d\omega_y=\frac{\mathcal{A}_y\,\lvert\cos\theta_y\rvert}{\lVert\mathbf{X}_y\rVert^2},
\]
and solving it for $\mathcal{A}_y$ gives \cref{eq:pixel-area}.
We keep $\lvert\cos\theta_y\rvert$ above $0.1$, so that a surface seen edge-on does not get an unbounded area.
\paragraph{Intensity and gain.}
A window's brightness is read from the photograph and corrected by one constant.
A window's intensity $\Phi_j=\bar Y_j$ (\cref{sec:windows}) already carries the camera's exposure, so windows are expressed
in the photograph's own units; with lamp intensities learned, the photograph's overall exposure needs no separate parameter.
The gain $\beta=20$ of \cref{eq:window-direct} is a fixed constant obtained empirically (\cref{supp:ablations}); it compensates
for the photograph recording a window darker than it is, because its pixels clip at white and are tone-mapped.
Depth is median-normalised, so the areas above are relative within an image; the window's absolute scale is carried by its
intensity and the gain $\beta$.

\subsection{Interreflection}
\label{supp:interrefl}

Interreflection (\cref{sec:interreflection}) needs two quantities from the network: how much light the room bounces, and where that light lands. How much is set by the effective mean albedo, $\bar\rho=0.3+0.2\,\sigma(\mathrm{Linear}(\mathbf{z}))$, with $\mathbf{z}$ the room code, globally pooled by the
extrinsic encoder.
The range $\bar\rho\in[0.3,0.5]$ of the effective mean albedo (\cref{sec:interreflection}) covers bedroom surfaces (walls $0.5$--$0.8$,
floors $0.2$--$0.4$, darker furniture and bedding).
Where it lands is set by the pattern, a normalised field,
\begin{equation}
  P_{i,x} = \frac{g_{i,x}}{\avg{g_i}_i},
  \qquad
  g_{i,x} = \operatorname{softplus}(a_{i,x}) + \SL\!\big(\mathbf{n}_x\!\cdot\mathbf{d}_{i,x}\big),
  \label{eq:field}
\end{equation}
where $\SL(t)=\operatorname{softplus}(1+t)/\operatorname{softplus}(1)$ is a smooth version of the Lambertian term $\pos{t}$, which we call a soft Lambertian: it equals $1$ where
$\mathbf{n}\perp\mathbf{d}$, grows towards $\mathbf{d}$ and never reaches zero, so $\mathbf{d}$ receives a gradient at every pixel, unlike $\pos{\mathbf{n}\cdot\mathbf{d}}$.
The field approximates indirect irradiance as a uniform term plus one dominant direction: $\operatorname{softplus}(a_{i,x})$ is bounce arriving from all directions, and $\SL(\mathbf{n}_x\!\cdot\mathbf{d}_{i,x})$ is bounce arriving mainly from $\mathbf{d}_{i,x}$, shaded by the surface normal.
This form is inspired by first-order spherical harmonics, but its coefficients vary per pixel and its normal term is the soft Lambertian, so it is a flexible local approximation rather than a spherical-harmonic lighting estimate.
Both fields of every light come from a small shared CNN with a learned position table that predicts a $256\times256$ grid.
Its inputs are the room code $\mathbf{z}$, the light-source masks, the normals and the position table; the photograph itself is not an input, so the fields see it only through $\mathbf{z}$ and cannot copy its texture.

\subsection{Rooms without an active light}
\label{supp:fallback}
A photograph with no light visible or on still needs a shading. It is the per-scene shading of \cref{eq:cast} (\cref{sec:scene}),
\begin{equation}
  s^{\mathrm{scene}}_{x} = c^{\mathrm{scene}}\, \SL\!\big(\mathbf{n}_x\!\cdot\mathbf{d}^{\mathrm{scene}}\big)\, B_x,
  \label{eq:fallback}
\end{equation}
a hemisphere light with one colour $c^{\mathrm{scene}}$ and one direction
$\mathbf{d}^{\mathrm{scene}}$ per image, both linear in $\mathbf{z}$ (the colour through a softplus), times a darkening map $B\in(0,1]$ predicted from the photo, normals
and depth at low resolution and upsampled along depth and normal edges.
The map $B$ gives the per-scene shading its spatial structure.
The per-scene shading is trained with the same shading target and loss as the per-light model, on photos with no light on and on
photos whose light-source masks are withheld, which we do for a random $20\%$ of the training photographs.
The fallback is rare: it is used for $1.5\%$ of LSUN validation photographs ($108$ of $7{,}398$) and for none of the $59$ paired photographs.
\Cref{fig:supp-fallback} shows three examples.
On SAW (\cref{tab:saw}), the $480$ test photographs without a detected light, which are shaded this way, score a challenge AP of $94.54$, against $94.76$ for the $1{,}219$ with one.

\subsection{Ray-traced visibility}
\label{supp:visibility}

A light casts a shadow where a surface blocks it (\cref{sec:visibility}), so we test whether each light can see each pixel by casting rays against a mesh built from the depth map. The mesh has two triangles per pixel quad and leaves out quads whose corners differ in depth by more than 6\%, so no surface
is stretched across a depth discontinuity.
A single depth map gives only the surfaces the camera sees: what lies behind an object is unknown and stays empty in the mesh, so a ray that passes there is not blocked, and shadows are cast by the visible surfaces only.
Each light $i$ is sampled at 64 points $\mathbf{y}_{i,s}$: a lamp's are drawn from its mask and lifted by the depth, and a
window's lie on the plane fitted to the wall around its mask, since the depth on window glass measures the outdoors.
The visibility of light $i$ at pixel $x$ is
\begin{equation}
  V_{i,x} = \frac{\sum_s w_{i,s,x}\,(1-b_{i,s,x})}{\sum_s w_{i,s,x}}, \qquad w_{i,s,x} = \lVert\mathbf{y}_{i,s}-\mathbf{X}_x\rVert^{-2},
  \label{eq:visibility}
\end{equation}
where $b_{i,s,x}=1$ if the segment from $\mathbf{X}_x$ to $\mathbf{y}_{i,s}$ hits the mesh.
A light's own pixels, dilated by 3 pixels, are removed from the mesh so that a shade does not shadow itself, and $V_i=1$ on them.
\Cref{fig:supp-shadows} shows edits without and with this visibility.

\subsection{The albedo}
\label{supp:albedo}

Every relit image multiplies the edited shading by a fixed albedo $A$, computed once per photograph (\cref{sec:editing}).
The product is formed in linear RGB and converted to sRGB for display, with values above 1 clipped.
The input is $A_{\mathrm{in}}$, the albedo that a pretrained intrinsic model~\cite{careaga2024colorful} predicts for the photograph; the output is the albedo $A$ we render with.
We work on its logarithm, $a=\log\mathrm{clip}(A_{\mathrm{in}},10^{-3},1)$.
$A_{\mathrm{in}}$ is accurate but keeps some shading (soft shadows, light falloff), which varies slowly across the image.
A surface's albedo is nearly constant, while leaked shading drifts slowly across it. So where the photograph is smooth we want the albedo to be flat, and we let it change only across an edge of the photograph.
We impose this with a photo-guided weighted total variation~\cite{rudin1992tv} in log space,
\begin{equation}
  b^{*}=\arg\min_{b}\;\tfrac12\sum_x\lVert b_x-a_x\rVert^2+\lambda\sum_{x\sim y}w_{xy}\lVert b_x-b_y\rVert,
  \label{eq:supp-wtv}
\end{equation}
where $b$ is the log albedo we solve for and $b^{*}$ its solution, $a$ the log of the input albedo, and $x$, $y$ index pixels: the first sum runs over all pixels and the second over pairs of
4-neighbours $x\sim y$. The norm is taken jointly over the three colour channels, $\lambda=15$, and the weights are
$w_{xy}=\exp(-\lVert q_x-q_y\rVert^2/0.15^2)$ with $q=\log(I_{\mathrm{lin}}+0.02)$, where $I_{\mathrm{lin}}$ is the photograph in linear RGB.
The weights carry the photograph's guidance: in its smooth regions $w_{xy}\approx1$, so differences between neighbouring albedo values are penalised and the slow variations, which are mostly leaked shading, are flattened; across one of its edges $w_{xy}\approx0$, so the albedo is free to change there and follows the input.
We solve \cref{eq:supp-wtv} with Chambolle--Pock~\cite{chambolle2011primal} ($1{,}500$ iterations, step sizes
$\tau=\sigma=0.99/\sqrt8$).
Total variation also flattens fine texture that is real albedo (fabric, wood grain), so texture finer than about 2 pixels
is restored from $A_{\mathrm{in}}$:
\begin{equation}
  A=\mathrm{clip}\big(\exp\!\big(G_2*b^{*}+a-G_2*a\big),\,10^{-3},\,1\big),
\end{equation}
where $G_2$ is a Gaussian of $\sigma=2$ pixels: the coarse albedo comes from $b^*$ and the fine detail from $A_{\mathrm{in}}$, where leaked
shading is negligible.
\Cref{fig:supp-albedo} shows the input and the filtered albedo of two photographs.

{\renewcommand{\thefigure}{B}\renewcommand{\theHfigure}{suppB}%
\begin{figure}[t]
  \centering
  \includegraphics[width=\linewidth]{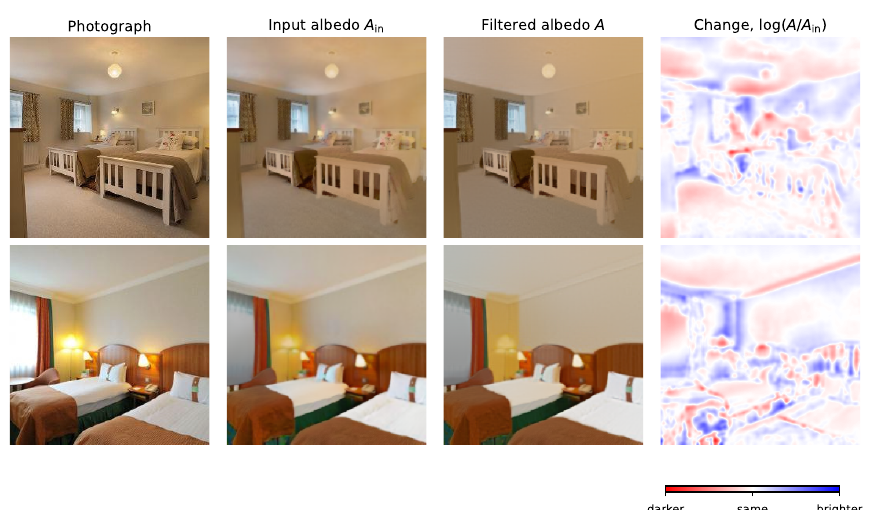}
  \caption{\textbf{The albedo filter} (\cref{supp:albedo}). The input albedo $A_{\mathrm{in}}$ keeps some slowly varying shading, such as the uneven light across the ceiling and the walls (top)
  and the glow of the lamps on the wall (bottom); the filtered albedo $A$ reduces it where the photograph is smooth and keeps the photograph's edges.
  Right: $\log(A/A_{\mathrm{in}})$, red where the filter darkened the albedo and blue where it brightened it.}
  \label{fig:supp-albedo}
\end{figure}}%
\addtocounter{figure}{-1}%

\section*{Training details and design choices}

\subsection{On/off pseudo-labels}
\label{supp:onoff}

Since our method is built on unlabelled single images, the on/off head (\cref{sec:lamps,sec:training}) is supervised with pseudo-labels computed from the photograph's luminance,
$Y=0.2126\,R+0.7152\,G+0.0722\,B$ (the Rec.~709 weights, applied to the sRGB values in $[0,1]$); no human annotation is used.
For a lamp with a shade we take four quantities from its mask:
the \emph{shade} mean $Y_s$ and its 95th percentile $Y_{95}$ over the mask rows above the shade's lower edge (the first row
below the widest row where the mask is less than half as wide);
the \emph{below} mean $Y_b$ over the pixels under that edge inside the mask's bounding box but outside the mask;
and the \emph{surround} mean $Y_r$ over the bounding box grown by its own size on every side, minus the mask.
The lamp is labelled
\begin{itemize}
  \item \textbf{off} if $Y_s/Y_r<0.7$, $Y_{95}<0.7$, $Y_b<0.5$ and $Y_b/Y_r<1$ (a dark shade with no light below it);
  \item otherwise \textbf{on} if $Y_{95}\ge0.93$ or $Y_s/Y_r\ge1.28$ (a blown-out or clearly glowing shade);
  \item otherwise \textbf{off} if $Y_b/Y_r<0.7$;
  \item otherwise it is left unlabelled and receives no on/off loss.
\end{itemize}
A ceiling fixture has no shade; it is labelled on if the 95th percentile of $Y$ over its whole mask
is at least $0.9$, and off otherwise.
Masks with fewer than 4 pixels are not labelled.
The thresholds were chosen empirically, once, by inspecting photographs, and are the same for every image.

\subsection{Losses and optimisation}
\label{supp:s1train}

Training (\cref{sec:training}) fits the summed shading to its target and adds three terms for what that target cannot teach: a lamp's on/off state, a window's own brightness, and how a lamp looks when off. The total loss is
\begin{equation}
  \mathcal{L} = \mathcal{L}_{S} + 3\,\mathcal{L}_{o} + 0.2\,\mathcal{L}_{W} + 0.2\,\mathcal{L}_{\mathrm{off}},
  \label{eq:loss}
\end{equation}
with $\mathcal{L}_{S}$ the shading loss of \cref{eq:loss-main}.
Both shadings are floored at $0.01$ before the logarithm, so that near-black pixels, where the logarithm is unstable and the
shading target least reliable, do not dominate the loss.

The shading target is not reliable on window pixels: a window shows the outdoors, not a surface of the room, and the intrinsic model cannot separate that view from shading (\cref{sec:training}).
On window pixels we therefore supervise from the photograph instead, and require the window mask's brightness relative to the room to match it,
$\mathcal{L}_{W}=\avg{\,\lvert\log(S_x/\avg{S}_{\mathrm{room}})-\log(Y_x/\avg{Y}_{\mathrm{room}})\rvert\,}_{\mathrm{win}}$, averaged over the window pixels, with
$Y$ the photo's linear luminance; where the photo clips, only a too-dark window mask is penalised, and a small term keeps the
window mask's colour close to grey.
Since off lamps are a minority in our training photographs, we train the on/off head with a focal binary cross-entropy~\cite{lin2017focal} (focusing parameter 2) on the labelled lamps,
$\mathcal{L}_{o}$, and $o_i$ is binarised with a straight-through estimator.
Finally, no training photograph shows what a lamp that is on looks like once it is switched off, so nothing supervises that state directly, and a switched-off fixture can stand out from its surroundings, brighter or darker.
An off-state term $\mathcal{L}_{\mathrm{off}}$ therefore re-renders a few images with every lamp switched off and
penalises a fixture that stays brighter or darker than a ring around it by more than $\lvert\log\text{ratio}\rvert=0.2$:
an off lamp should look like the surfaces next to it.
We train for one epoch over the $83$k photographs ($5.2$k iterations at batch size $16$). This takes about $2$ GPU-hours on one NVIDIA A40, after the frozen pretrained models have computed the inputs and targets once.
We optimise with AdamW, learning rate
$1.5\cdot10^{-4}$ ($10\times$ lower for the extrinsic encoder, $16\times$ higher for the $17$ scalar gains and blend weights, $\gamma$ among them, and $14\times$ higher for the per-scene shading's colour and direction), weight decay $10^{-4}$, gradient clipping at $1$ and horizontal-flip augmentation.
\Cref{tab:supp-arch} lists the model's components and \cref{tab:supp-cost} the training and inference cost.

{\renewcommand{\thetable}{A}\renewcommand{\theHtable}{suppA}%
\begin{table}[t]
\caption{\textbf{Interreflection and shadows.} On the $128$ test pairs, tuned and scored as in \cref{tab:ablation}. Each part is removed at inference, except in the
retrained row, which is the ``w/o interreflection'' row of \cref{tab:ablation}. \emph{Unlit}: the pixels whose shading is below $2\%$ of the image's median, in the untuned edit.
Bold: best per column.}
\label{tab:supp-abl4}
\centering
\footnotesize
\setlength{\tabcolsep}{3pt}
\begin{tabular}{@{}lcccc@{}}
\toprule
Variant & Unlit & PSNR$\uparrow$ & SSIM$\uparrow$ & LPIPS$\downarrow$ \\
\midrule
Ours                               & $0\%$  & \textbf{18.66} & \textbf{0.734} & 0.255 \\
w/o ray-traced shadows             & $0\%$  & 18.28          & \textbf{0.734} & 0.255 \\
w/o interreflection                & $12\%$ & 15.76          & 0.597          & 0.366 \\
w/o interreflection, retrained     & $12\%$ & 15.71          & 0.597          & 0.368 \\
w/o interreflection and shadows    & $9\%$  & 16.04          & 0.651          & 0.320 \\
Interreflection ignores the edit   & $0\%$  & 18.05          & 0.713          & \textbf{0.248} \\
\bottomrule
\end{tabular}
\end{table}}%
\addtocounter{table}{-1}%
{\renewcommand{\thetable}{B}\renewcommand{\theHtable}{suppB}%
\begin{table*}[t]
\caption{\textbf{What the ray-traced shadows change.} On the $128$ test pairs, tuned and scored as in \cref{tab:ablation}. \emph{Shadow region}: the pixels the shadows darken by more
than $15\%$ in our untuned edit, $17\%$ of the image in the median pair; \emph{rest}: all other pixels. Within a region, SSIM and LPIPS are the means of their per-pixel maps
(for LPIPS, of its spatial variant), leaving out a $5$-pixel image border as SSIM does. The last four rows change only the visibility maps $V_i$ of \cref{supp:visibility}:
smoothed with a Gaussian of standard deviation $\sigma$, or moved $4$ pixels sideways.}
\label{tab:supp-shadow}
\centering
\footnotesize
\begin{tabular}{@{}lccccccccccc@{}}
\toprule
& \multicolumn{3}{c}{Whole image} && \multicolumn{3}{c}{Shadow region} && \multicolumn{3}{c}{Rest of the image} \\
\cmidrule{2-4}\cmidrule{6-8}\cmidrule{10-12}
Shadows & PSNR$\uparrow$ & SSIM$\uparrow$ & LPIPS$\downarrow$ && PSNR$\uparrow$ & SSIM$\uparrow$ & LPIPS$\downarrow$ && PSNR$\uparrow$ & SSIM$\uparrow$ & LPIPS$\downarrow$ \\
\midrule
None                           & 18.28 & 0.734 & 0.255 && 16.23 & 0.614 & 0.267 && 18.81 & 0.756 & 0.255 \\
Ray-traced (ours)              & 18.66 & 0.734 & 0.255 && 17.75 & 0.627 & 0.253 && 18.91 & 0.754 & 0.258 \\
\midrule
Smoothed, $\sigma=2$ pixels    & 18.69 & 0.740 & 0.252 && 17.79 & 0.646 & 0.251 && 18.92 & 0.757 & 0.255 \\
Smoothed, $\sigma=4$ pixels    & 18.68 & 0.740 & 0.253 && 17.70 & 0.646 & 0.254 && 18.91 & 0.757 & 0.255 \\
Smoothed, $\sigma=8$ pixels    & 18.63 & 0.738 & 0.254 && 17.50 & 0.641 & 0.258 && 18.88 & 0.756 & 0.256 \\
Moved by $4$ pixels            & 18.61 & 0.728 & 0.259 && 17.56 & 0.615 & 0.257 && 18.86 & 0.749 & 0.262 \\
\bottomrule
\end{tabular}
\end{table*}}%
\addtocounter{table}{-1}%
{\renewcommand{\thefigure}{C}\renewcommand{\theHfigure}{suppC}%
\begin{figure*}[t]
  \centering
  \includegraphics[width=\linewidth]{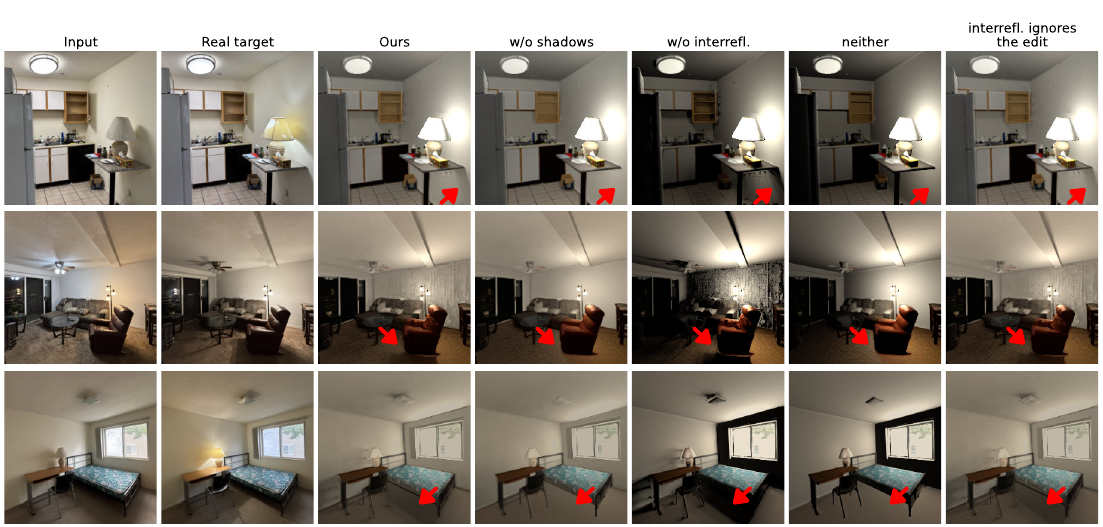}
  \caption{\textbf{Ablation variants} (\cref{supp:ambient}, \cref{tab:supp-abl4}) on three test pairs, untuned: every light keeps the colour and intensity our model estimates, so the variants differ only in the part removed
  (\cref{tab:supp-abl4} scores them tuned, which also refits each light's colour). Each image is shown after one gain per colour channel fitted to the target, as scored. Red arrows: where the ray-traced shadows fall, at the same place in every edit.
  Without interreflection, every surface a light does not see directly turns black; without shadows, the lamp and the ceiling light reach surfaces hidden from them.}
  \label{fig:supp-abl4}
\end{figure*}}%
\addtocounter{figure}{-1}%

\subsection{Ablations of shadows and interreflection}
\label{supp:ambient}
The ablation (\cref{sec:comparisons}, \cref{tab:ablation}) removes the interreflection or the ray-traced shadows.
\Cref{tab:supp-abl4} adds three variants, \cref{fig:supp-abl4} shows them on three test pairs, and \cref{tab:supp-shadow} examines what the shadows change.
Every variant is tuned and scored as in \cref{tab:ablation}.
In short, interreflection is essential, each light's interreflection has to follow its edit, and the shadows improve every metric where they fall.

\paragraph{Interreflection is essential.}
Without interreflection, a surface that no light reaches directly receives no light: $12\%$ of the pixels are left unlit, against none in our edits
(\cref{tab:supp-abl4}; \cref{fig:supp-abl4}, fifth column), and PSNR falls by $2.9$\,dB.
Removing the interreflection at inference and retraining the model without it, as in \cref{tab:ablation}, score almost the same ($15.76$ and $15.71$\,dB, SSIM $0.597$ for both), and for a reason.
Untuned, the retrained model is the better one ($12.82$ against $11.98$\,dB): it learns lights about twice as bright to make up for the missing interreflection.
Tuning then fits the intensity and colour of every light to the target in both models, which removes that difference; what is left is the same physical model of direct light,
with the same $12\%$ of the pixels unlit.

\paragraph{Interreflection follows the edit.}
An edit changes a light's whole layer: its emission, its direct light and its interreflection (\cref{eq:formation}).
To test this tie, we edit only the emission and the direct light and keep each light's interreflection at the level it has in the input photograph, as an ambient term would
(\cref{tab:supp-abl4}, last row).
PSNR falls by $0.6$\,dB and SSIM by $0.02$. LPIPS slightly prefers this variant ($0.248$ against $0.255$), because it leaves more of the photograph as it was (\cref{sec:comparisons}).
We keep the tie: a lamp that is switched off cannot light the room by bounce.

\paragraph{Shadows need interreflection.}
Without interreflection, removing the shadows as well raises the scores ($16.04$ against $15.76$\,dB, SSIM $0.651$ against $0.597$).
The edit without interreflection is already too dark, and shadows only take light away: with them, the unlit pixels grow from $9\%$ to $12\%$.
A shadow is right only when interreflection fills it in, which is why our model has both. Either way, the edit without interreflection stays more than $2.6$\,dB below ours.

\paragraph{Shadows improve every metric where they fall.}
In the part of the image the ray-traced shadows of \cref{supp:visibility} act on, they improve PSNR, SSIM and LPIPS alike (\cref{tab:supp-shadow}).
This \emph{shadow region}, the pixels the shadows darken by more than $15\%$, makes up $17\%$ of the image in the median pair; inside it the shadows raise PSNR by $1.5$\,dB and
SSIM by $0.013$ and lower LPIPS by $0.015$, and over the whole image they raise PSNR by $0.38$\,dB, in $88\%$ of the pairs.
The whole-image SSIM and LPIPS of \cref{tab:ablation} stay the same only because a small loss on the rest of the image cancels this gain.
That loss comes from pixel-scale detail in the traced shadow maps, not from where the shadows are: once the detail is smoothed away, the shadows win on all three metrics over the whole image.

The tie in whole-image SSIM and LPIPS is a cancellation.
Outside the shadow region, SSIM falls by $0.002$ and LPIPS rises by $0.003$; the gain on a sixth of the pixels and this small loss on the rest cancel in the mean.

The loss is pixel-scale detail, and smoothing removes it.
The traced visibility has crisp edges and speckle, left by a mesh built on a single-view depth map at $256\times256$, and SSIM and LPIPS,
which compare local structure, count this detail against the edit wherever the target does not share it exactly.
Smoothing each visibility map $V_i$ with a Gaussian of $\sigma=2$ pixels leaves every shadow in place, removes the loss outside the shadow region and more than doubles the SSIM gain inside it.
Over the whole image the edit then beats the one without shadows on all three metrics ($18.69$\,dB, $0.740$, $0.252$), in $90\%$, $76\%$ and $65\%$ of the pairs; $\sigma=4$ and $8$ give similar results.
The cost of the detail grows with the amount of shadow: on the quarter of the pairs with the largest shadow region, the shadows as traced take SSIM from
$0.727$ to $0.719$ and LPIPS from $0.269$ to $0.285$, and smoothed they give $0.730$ and $0.273$.

The shadows are well placed.
Moving the visibility maps $4$ pixels sideways, with their edges as crisp, keeps most of the PSNR gain but takes SSIM and LPIPS below the edit without shadows ($0.728$, $0.259$).
These two metrics respond to where exactly a shadow's edge is, and our shadows beat the moved ones in $84\%$, $88\%$ and $78\%$ of the pairs.

All our results use the shadows as traced.
The smoothing was found on the test pairs, so we do not use it anywhere else in this paper. \Cref{fig:supp-shadows} shows the shadows on two pairs.

\begin{table}[t]
\caption{\textbf{Ours and Li \etal, tuned and untuned.} On the $128$ test pairs, scored and averaged as in \cref{sec:setup}.
Tuned: one RGB gain per light fitted to the target (for every light, or only for the lights the edit switches on), switched-off lights kept off. Bold, underlined: best, second per column; italic: the unedited photograph, for reference.}
\label{tab:supp-tuning}
\centering
\footnotesize
\setlength{\tabcolsep}{3pt}
\begin{tabular}{@{}lcccc@{}}
\toprule
Method & PSNR$\uparrow$ & SSIM$\uparrow$ & LPIPS$\downarrow$ & \hcell{Beats\\unedited} \\
\midrule
Ours, untuned                      & 16.90             & 0.696             & 0.258             & \underline{13\%} \\
Ours, tuned for PSNR (main)    & \underline{18.66} & \textbf{0.734}    & \underline{0.255} & 12\%             \\
Ours, tuned for LPIPS                & 17.94             & 0.729             & \textbf{0.237}    & \textbf{14\%}    \\
Ours, switched-on lights only        & 17.83             & 0.717             & 0.258             & 12\%             \\
\midrule
Li \etal, untuned                  & 17.49             & 0.710             & 0.295             & 0\%              \\
Li \etal, tuned for PSNR (main) & \textbf{18.99}    & \underline{0.731} & 0.308             & 0\%              \\
Li \etal, tuned for LPIPS            & 18.15             & 0.722             & 0.267             & 2\%              \\
Li \etal, switched-on lights only    & 18.13             & 0.717             & 0.308             & 0\%              \\
\midrule
\textit{Unedited photo}              & \textit{18.57}    & \textit{0.813}    & \textit{0.159}    & --               \\
\bottomrule
\end{tabular}

\end{table}

\subsection{Fixed parameters}
\label{supp:ablations}

Two parameters of the light models (\cref{sec:lamps,sec:windows}) are fixed because, when learned, they absorbed brightness instead of describing the light source.
\paragraph{Fixed cone width.}
The exponent $\kappa_i$ of \cref{eq:lobe} sets how narrow a lamp's cones are.
When it was learned, it drifted to compensate for lamp brightness (collapsing to $0.1$ or growing into a needle) instead of describing the fixture.
We therefore fix it by lamp type: $\kappa_i=2$ for ceiling fixtures, a wide cone, and $\kappa_i=8$ for other lamps, a narrow one.

\paragraph{Fixed window gain.}
A window's intensity is its mean pixel value (\cref{sec:windows}), which understates the light it casts, because a window's pixels clip at white and are tone-mapped.
The gain $\beta$ of \cref{eq:window-direct} corrects for this.
We fix it to $\beta=20$ for every window rather than predict it per window: a learned gain gave the darkest windows the largest value, so a dark night
window still cast most of the room's light. A window's light reaches most of the room, so a free gain lets it explain light that it does not emit.
The value $20$ is the median gain learned on windows brighter than their room.

\section*{Evaluation on the paired photographs}

\subsection{Tuned and untuned}
\label{supp:tuning}
A light that is off in the input photograph shows nothing of its brightness, so a method with explicit lights has to be given it.
Following LuxRemix~\cite{liang2026luxremix}, which tunes each method's light-editing parameters to each target photograph, \cref{sec:setup} and \cref{tab:vs-unedited} report ours and Li \etal tuned for PSNR.
Fitting for PSNR makes some of Li \etal's edits look worse, so \cref{fig:qual} shows Li \etal as published, its better-looking edits. \Cref{tab:supp-tuning} gives both methods
without tuning, tuned for PSNR as in \cref{sec:setup}, and tuned for LPIPS, always with every light the edit switches
off kept off. Tuning for PSNR raises Li \etal's PSNR by $1.5$\,dB but worsens its LPIPS. With both tuned for PSNR, as in \cref{tab:vs-unedited}, ours has the lower LPIPS on $77\%$ of the pairs. Under every combination, ours has
the lower mean LPIPS, and Li \etal comes closer to the target than the unedited photograph on at most $2\%$ of pairs.
Both hold when only the lights the edit switches on, whose brightness the input cannot show, are fitted.

\subsection{More comparisons on the paired photographs}
\label{supp:gain}
\begin{figure}[t]
  \centering
  \IfFileExists{figs/scatter_lpips.pdf}{\includegraphics[width=0.8\linewidth]{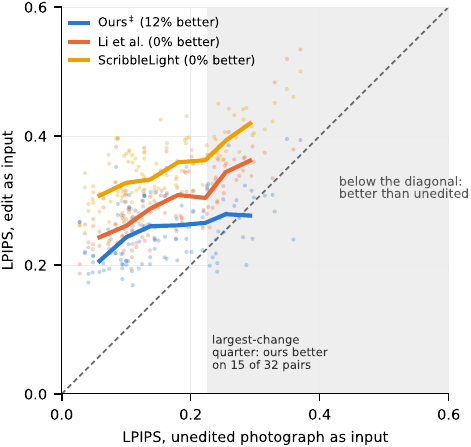}}{\fbox{\parbox[c][4cm][c]{0.9\linewidth}{\centering LPIPS scatter}}}
  \caption{\textbf{Every pair under LPIPS} (lower is better). For each of the $128$ pairs, $x$ is the LPIPS of the unedited input photograph against the target and $y$ that of the method's edit.
  The larger $x$, the more the switch changes the room, and the worse leaving the photograph unedited scores. Below the dashed diagonal, the
  edit is closer to the target than the unedited photograph; each thick line is that method's mean in bins of $x$.
  Shaded: the quarter of pairs where the switch changes the room most. There ours is closer to the target than the unedited photograph on $15$ of $32$ pairs, and Li \etal and ScribbleLight on none: of the three methods plotted, only ours beats the unedited photograph on this metric as well as in the user study.
  $^\ddagger$Light intensities fitted to the target; Li \etal as published. LumiNet,
  which sees the target, is not plotted. LightLab~\cite{lightlab} and LuxRemix~\cite{liang2026luxremix} do not release their models, so they cannot be run on these pairs.}
  \label{fig:scatter-lpips}
\end{figure}

Beyond the comparisons of \cref{sec:comparisons}, we showcase an extra eight pairs drawn at random from the $128$ test pairs (\cref{fig:random-pairs}). We also plot every pair under LPIPS (\cref{fig:scatter-lpips}), show Qwen-Image-Edit and ScribbleLight hallucinating, removing or replacing light fixtures (\cref{fig:qwen}), and give the gain of each method over the unedited photograph, pair by pair (\cref{fig:gain-curves}).
{\renewcommand{\thefigure}{D}\renewcommand{\theHfigure}{suppD}%
\begin{figure}[t]
  \centering
  \includegraphics[width=\linewidth]{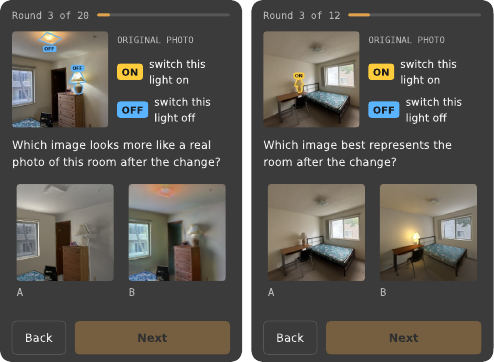}
  \caption{\textbf{One round of the user study under each question} (\cref{supp:userstudy}), redrawn from the study page as shown on a phone.
  Each round shows the input with the lights to switch marked, the question and two candidates in random order.
  Left: our edit (A) against Li \etal's (B). Right: the unedited photograph (A) against our edit (B).}
  \label{fig:supp-userstudy}
\end{figure}}%
\addtocounter{figure}{-1}%

The \emph{Fixtures intact} labels of \cref{tab:vs-unedited} are annotated by a human, one pair at a time: every method's edit of each pair is viewed
beside its input and real target, with close-ups where needed, and the pair is marked as altered when a light fixture is removed,
replaced by a different one or added where the target has none; a change in brightness or colour alone does not count.
We will release the per-pair labels with the dataset.

\paragraph{Changed and unchanged regions.}
Whole-image metrics mix two things: how well an edit makes the requested change, and how much it disturbs what should stay.
\Cref{tab:supp-region} separates them, scoring each output inside the quarter of its pixels where the unedited photograph differs most from
the target, and on the rest.
In the changed region, where an edit has to make the requested change, ours makes it best of the methods that keep the light fixtures intact: it is $3.2$\,dB closer to the target than the unedited photograph ($16.72$ against $13.48$), ahead of Li \etal ($16.37$) and ScribbleLight ($14.60$).
On the rest of the image, where little should change, ours is $5.3$\,dB below the unedited photograph ($19.87$ against $25.13$).
This loss comes from re-rendering: every edited image is the albedo times the edited shading, which does not reproduce the photograph's fine detail even where the light does not change (\cref{sec:limitations}).
Qwen-Image-Edit scores higher than ours in both regions ($16.93$ and $24.23$), because it largely reproduces the photograph; but it hallucinates, removes or replaces a light fixture in $73\%$ of pairs (\cref{fig:qwen}), and a text prompt cannot set a light's intensity or colour as precisely as the parameters of a physical light model.
The pattern holds for the tenth and the half of the pixels that change most (ours $15.6$ and $17.6$\,dB, unedited $10.8$ and $15.8$\,dB).

\begin{figure*}[t]
  \centering
  \includegraphics[width=\linewidth]{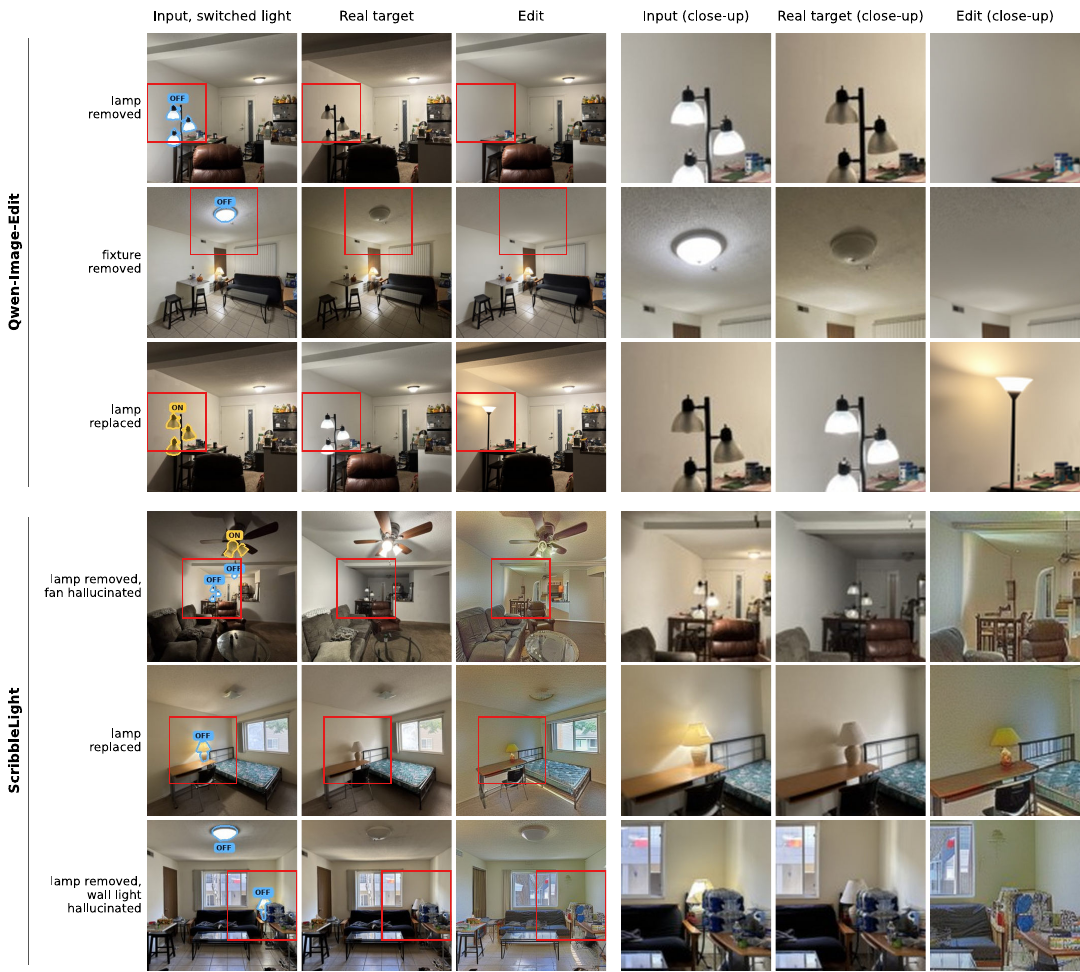}
  \caption{\textbf{Editing the scene instead of its light} (\emph{Fixtures intact} in \cref{tab:vs-unedited}). Each row: the input with the
  switched light marked, the real target and the edit, then close-ups of the boxed region. Asked to switch a light,
  Qwen-Image-Edit~\cite{wu2025qwenimage} removes the floor lamp (row 1) or the ceiling fixture (row 2), or replaces the floor lamp
  with another lamp (row 3); ScribbleLight~\cite{scribblelight} removes the floor lamp and hallucinates a ceiling fan in the hallway (row 4),
  replaces the table lamp (row 5), and removes the table lamp and hallucinates a wall light (row 6). The real target keeps every fixture.}
  \label{fig:qwen}\label{fig:fixtures}
\end{figure*}

\begin{figure*}[t]
  \centering
  \IfFileExists{figs/improvement_curves.pdf}{\includegraphics[width=\linewidth]{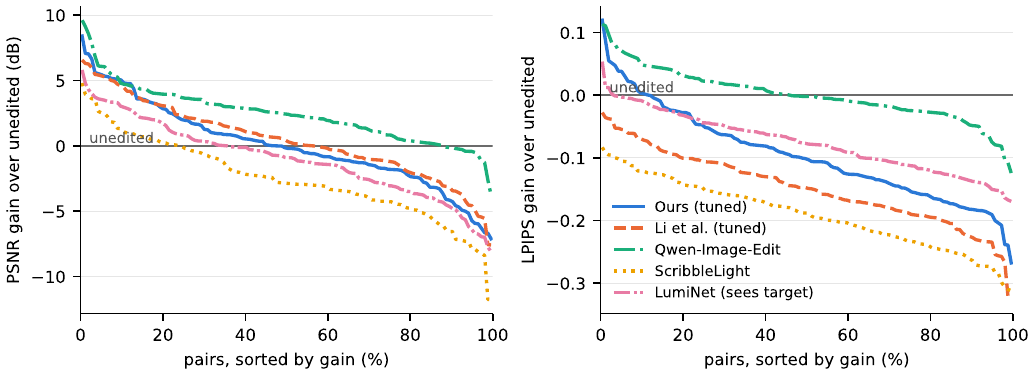}}{\fbox{\parbox[c][4cm][c]{0.9\linewidth}{\centering gain curves}}}
  \caption{\textbf{Gain over the unedited photograph, pair by pair.} For each method, the gain of its edit over leaving the
  photograph unedited on each of the $128$ pairs, sorted from largest to smallest: PSNR (left, dB) and LPIPS (right, unedited
  minus edit, so higher is better). Above zero, the edit is closer to the target than the input it started from. Ours, trained on single images, is level with Li \etal, trained on multi-light data, on PSNR ($18.66$ against $18.99$\,dB on average) and ahead on LPIPS ($0.255$ against $0.308$).
  In the user study, participants preferred ours to Li \etal's in $84\%$ of comparisons (\cref{supp:userstudy}), which is the order LPIPS gives.
  Ours and Li \etal are tuned (\cref{sec:setup}).
  LumiNet receives the target photograph as its lighting reference, and Qwen-Image-Edit hallucinates, removes or replaces a light fixture in $73\%$ of pairs (\cref{fig:qwen}).}
  \label{fig:gain-curves}
\end{figure*}

\subsection{User study}
\label{supp:userstudy}
In the user study (\cref{sec:comparisons}), each round showed the input photograph with the lights to be switched marked on or off, and two candidate images side by side in random order; the participant picked one (\cref{fig:supp-userstudy}).
Rounds were drawn from the $128$ test pairs, balanced over the $19$ scenes. Our edits and Li \etal's are tuned to the target.
\Cref{tab:supp-userstudy} gives the counts of every comparison.

\paragraph{Against other methods.}
The question was ``Which image looks more like a real photo of this room after the change?''.
$33$ participants made $531$ choices between two of our edit, Li \etal's edit and the real target photograph.
They preferred ours to Li \etal's in $84\%$ of comparisons, and picked the real photograph over ours in $95\%$ and over Li \etal's in $98\%$.
$10$ further participants made $118$ choices between our edit and ScribbleLight's, shown as scored (\cref{sec:setup}), and preferred ours in $93\%$.

\paragraph{Against the unedited photograph.}
The question was ``Which image best represents the room after the change?''.
$13$ participants made $151$ choices between our edit and the input photograph itself.
They preferred our edit in $92\%$, although LPIPS ranked the unedited photograph closer to the target in $85\%$ of these rounds.

\paragraph{Significance.}
Every comparison differs from chance with binomial $p<10^{-23}$, and all participants but one preferred ours more often than the alternative
($31$ of $32$ against Li \etal, $10$ of $10$ against ScribbleLight, $13$ of $13$ against the unedited photograph).

\newpage
\section*{Evaluation on LuxRemix's and LightLab's data}

\subsection{Evaluation on the LuxRemix test set}
\label{supp:luxremix}
\begin{figure*}[t]
  \centering
  \includegraphics[width=\linewidth]{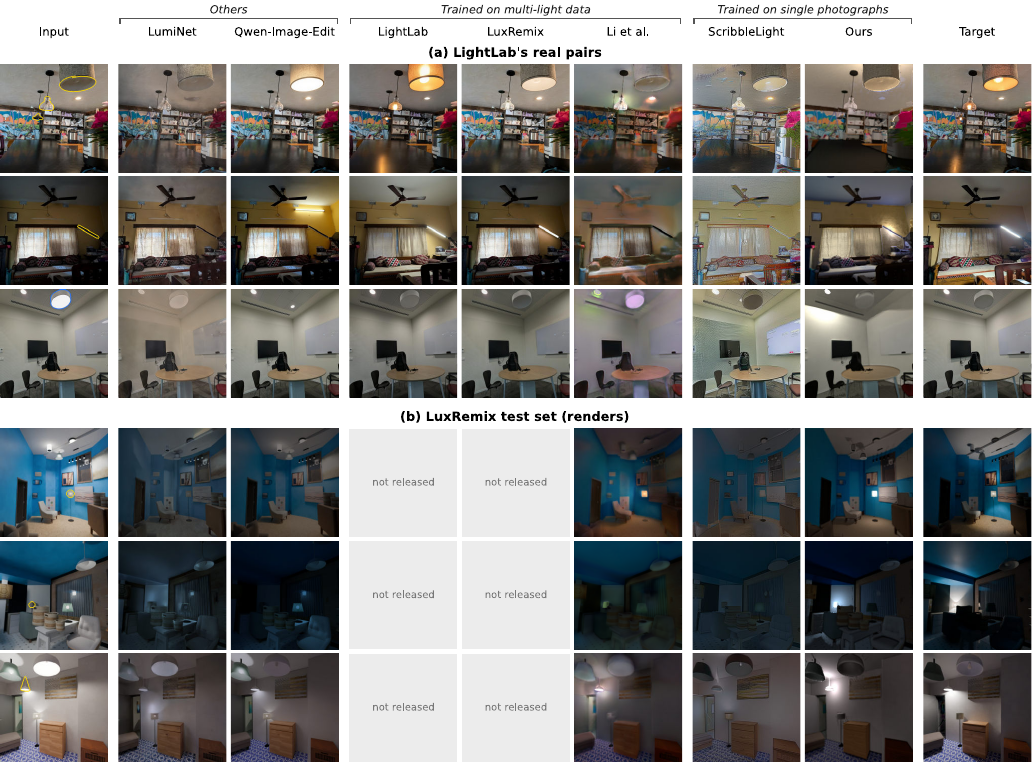}
  \caption{\textbf{LightLab and LuxRemix on their own data.} (a)~Three of LightLab's four eligible pairs. (b)~Three of
  LuxRemix's $112$ rendered test cases, the room lit by the outlined light alone; neither LuxRemix nor LightLab released outputs for
  them. LumiNet sees the target photograph, Qwen-Image-Edit receives a text prompt naming the light, ScribbleLight receives
  scribbles from the lights, and Li \etal and ours are tuned in (a); columns are grouped as in \cref{tab:lux-lightlab}, and every tile is shown as scored (in (b), by our protocol,
  one least-squares gain per colour channel).}
  \label{fig:supp-lux-lightlab}
\end{figure*}
\begin{table}[!tb]
\caption{\textbf{LuxRemix test set.} The image of a rendered room lit by one of its lights alone, from the render with every
light on ($112$ lights, $30$ rooms; means over lights). Top: our runs, scored by our protocol (\cref{sec:setup}) without tuning;
\emph{Unedited}: the input render, scaled per colour channel to the target. Bottom: $^*$reported by LuxRemix~\cite{liang2026luxremix},
whose scoring is not public, so the two blocks are not directly comparable; FLUX token, SD and LuxRemix-SV are trained on renders
from the same generator as the test rooms. $^\dagger$Sees the target; $^\P$removes or replaces fixtures; italic: the unedited input and the methods marked $^\dagger$ or $^\P$;
bold and underline: best and second within a block among the other rows.}
\label{tab:luxremix}
\centering
\small
\setlength{\tabcolsep}{4pt}
\begin{tabular}{@{}lccc@{}}
\toprule
Method & PSNR$\uparrow$ & SSIM$\uparrow$ & LPIPS$\downarrow$ \\
\midrule
\multicolumn{4}{@{}l}{\emph{Our runs, our protocol}} \\
\textit{Unedited render} & \textit{18.49} & \textit{0.769} & \textit{0.206} \\
\textit{LumiNet}$^\dagger$~\cite{xing2025luminet} & \textit{17.59} & \textit{0.727} & \textit{0.262} \\
\textit{Qwen-Image-Edit}$^\P$~\cite{wu2025qwenimage} & \textit{17.32} & \textit{0.742} & \textit{0.247} \\
ScribbleLight~\cite{scribblelight} & 16.76 & 0.570 & 0.451 \\
Li \etal~\cite{li2022physically} & \underline{16.85} & \underline{0.658} & \underline{0.382} \\
Ours & \textbf{17.46} & \textbf{0.691} & \textbf{0.328} \\
\midrule
\multicolumn{4}{@{}l}{\emph{Reported by LuxRemix}$^*$} \\
ScribbleLight~\cite{scribblelight} & 14.39 & 0.395 & 0.688 \\
Qwen-Image~\cite{wu2025qwenimage} & 18.23 & 0.714 & 0.237 \\
FLUX token~\cite{liang2026luxremix} & 25.20 & \underline{0.865} & 0.101 \\
SD~\cite{liang2026luxremix} & \underline{27.13} & 0.857 & \underline{0.099} \\
LuxRemix-SV~\cite{liang2026luxremix} & \textbf{27.68} & \textbf{0.898} & \textbf{0.082} \\
\bottomrule
\end{tabular}
\end{table}
\begin{table}[!tb]
\caption{\textbf{The LuxRemix case shown in its paper~\cite[Fig.~7]{liang2026luxremix}.} Scored by our protocol from the printed figure. Bold, underlined: best, second.}
\label{tab:supp-luxcase}
\centering
\small
\begin{tabular}{lccc}
\toprule
Method & PSNR$\uparrow$ & SSIM$\uparrow$ & LPIPS$\downarrow$ \\
\midrule
Ours & 18.83 & 0.719 & 0.335 \\
Qwen-Image & 19.23 & 0.747 & 0.245 \\
ScribbleLight & 20.01 & 0.767 & 0.279 \\
Li \etal & 20.84 & 0.780 & 0.290 \\
FLUX token & 26.33 & 0.886 & 0.134 \\
SD & \underline{28.11} & \underline{0.890} & \underline{0.118} \\
LuxRemix-SV & \textbf{28.80} & \textbf{0.913} & \textbf{0.101} \\
\bottomrule
\end{tabular}

\end{table}

\paragraph{Cases.}
The test set (\cref{sec:comparisons}, \cref{tab:lux-lightlab}) is LuxRemix's own list of evaluation cases~\cite{liang2026luxremix}: one view of each of $30$ rendered test
rooms and the $112$ lights it scores in these views; every score below is a mean over these $112$ lights.
We render the views with their released tools and camera parameters; each method receives the render with every light at
full power (tone-mapped, $512\times512$), and the target is the render of the same view lit by the scored light alone.
Our depth uses each view's own field of view ($61$--$97^\circ$) instead of the $55^\circ$ of \cref{supp:geometry}.

{\renewcommand{\thefigure}{E}\renewcommand{\theHfigure}{suppE}%
\begin{figure}[t]
  \centering
  \includegraphics[width=\linewidth]{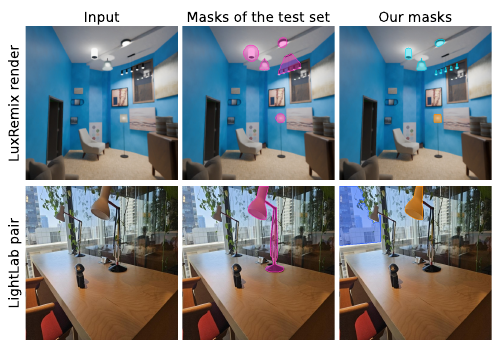}
  \caption{\textbf{The test sets' own masks and ours.} Top (\cref{supp:luxremix}): LuxRemix's masks (magenta) are prompts for its model: a disc around the
  cylinder light and the floor lamp, the convex hull of the pendant and of the whole track light. Ours (ceiling lights cyan, lamps orange, windows blue, as in
  \cref{fig:supp-masks}) outline each fixture, with one mask per track head. Bottom (\cref{supp:lightlab}): LightLab's mask is the switched lamp, with its arm and base;
  ours keeps the shade and adds the window.}
  \label{fig:supp-theirmasks}
\end{figure}}%
\addtocounter{figure}{-1}%
\paragraph{Light masks.}
Our model reads each light's position, extent and class from its mask, so on these renders it gets the kind of mask it was trained on:
SAM~3 on the input render with the prompts of \cref{supp:masks}, reviewed by hand as for the paired photographs.
In the review, duplicates of one fixture, reflections, highlights and unlit fixtures are removed, ceiling fixtures and pendants become ceiling lights,
a track light gets one mask per head, and two glowing signs that SAM~3 missed are drawn by hand.
LuxRemix's test set has a mask for each light too, but of another kind (\cref{fig:supp-theirmasks}): a prompt that tells LuxRemix's model which light is meant.
It is rendered from the scene's geometry and is the emitting source for $52$ of the $112$ lights, a small disc around it for $25$, the fixture for $9$ and
the fixture's convex hull for $26$; for half of the lights it overlaps our mask by less than half (intersection over union below $0.5$).
We use these masks only to match lights.
The choice matters little: with LuxRemix's masks as our model's input instead, ours scores $17.25$ / $0.679$ / $0.334$ (PSNR / SSIM / LPIPS), against
$17.46$ / $0.691$ / $0.328$ with ours (\cref{tab:luxremix}).

\paragraph{Matching lights.}
LuxRemix counts $112$ lights in the $30$ views and our masks hold $143$, because our model takes one mask per connected fixture:
a track light is one light to LuxRemix and one mask per head to our model.
Of LuxRemix's $112$ lights, $103$ correspond to one mask of ours each, $7$ to several masks each ($35$ masks for these $7$ lights), and $2$ to none.
Each of our masks is assigned to the LuxRemix light whose mask it overlaps most, or, if it overlaps none, to the nearest one
within $12$ pixels at $256\times256$ (a track head next to the mask of its track).
In one view, LuxRemix's masks of two overlapping pendants are nearly the same disc; there we assign the two lights by hand,
from LuxRemix's single-light renders.
The $2$ lights without a mask of ours (one outside the frame, one small fixture) are scored as a black image.
The remaining $5$ of our $143$ masks, two windows and three fixtures that match no LuxRemix light, are left out.
A LuxRemix light's predicted image is the sum, over the lights assigned to it, of each light's emission, direct light times
its ray-traced visibility and bounce, times our albedo, upsampled from $256$ to $512$ pixels.

\paragraph{Li \etal.}
We run their released pipeline with test-time optimisation and their own depth, with one lamp mask per light of ours
(masks too small for their lamp model are grown by $1$--$3$ pixels, $13$ lights).
A LuxRemix light's image is rendered by their renderer with only the lamps assigned to it switched on, keeping each lamp's
fitted intensity; their other lamps, windows and the invisible lamp and window are switched off.

\paragraph{Results.}
The paired photographs record each scene under a few combinations of its lights, never one light alone.
LuxRemix~\cite{liang2026luxremix} renders rooms one light at a time, and its single-image test set asks for the image of a
room lit by one of its lights alone, given the render with every light on: $112$ lights in $30$ rooms.
We run ours, Li \etal, ScribbleLight, LumiNet and Qwen-Image-Edit on these renders as in \cref{sec:experiments} and score every output
as in \cref{sec:setup}, without tuning (examples in \cref{fig:supp-lux-lightlab}b); the rows of \cref{tab:luxremix} marked $^*$ are reported by LuxRemix, whose scoring code
is not public, so they are indicative.
ScribbleLight and Qwen-Image-Edit appear in both blocks with different scores, because the blocks are different runs under different scoring:
ours use scribbles drawn from the light masks and a prompt that names the light to keep, and are rescaled to the target before scoring (\cref{sec:setup});
LuxRemix does not release the inputs it gave these methods, their outputs or its scoring code.
As on the paired photographs, leaving the render unedited is a strong baseline: scaled to the target, the render with every light
on is closer to the single-light render than ours on all but $7$ of the $112$ lights under LPIPS, and closer than Li \etal's on
all of them.
Ours is closer than Li \etal's on $85$ of the $112$ lights (LPIPS; Wilcoxon signed-rank $p<10^{-10}$).
The $30$ test rooms come from the held-out split of LuxRemix's own dataset ($400$ of its $12{,}439$ rendered scenes), so the test is in-distribution for the
methods trained on the rest of it (FLUX token, SD and LuxRemix-SV), which are far ahead of every other method (\cref{tab:luxremix}); ours has seen only photographs.
This is what images of the same room under different lights buy: where they can be rendered for the target domain, they should
be used; our model shows how far one gets without them.

\paragraph{The case shown by LuxRemix.}
LuxRemix released neither a model nor per-case results, so its rows in \cref{tab:luxremix} are quoted.
Its paper shows one of the $112$ cases for every method~\cite[Fig.~7]{liang2026luxremix}; scored by our protocol from that
figure, the methods rank as in \cref{tab:supp-luxcase} (their figure's own ground-truth panel scores $46.25$\,dB against the ground truth, the
loss of cropping a printed figure).
One case is not a benchmark, but LuxRemix-SV's scores on it are close to its means in \cref{tab:luxremix}.

\subsection{The LightLab pairs}
\label{supp:lightlab}
\begin{table}[!tb]
\caption{\textbf{LightLab's real photograph pairs.} The four in which a visible light stays on in both photographs, scored as LuxRemix
does: $512\times512$, display space, no rescaling. Top: outputs shared by LuxRemix's authors; bottom: our runs (ScribbleLight with
scribbles from the lights; ours and Li \etal tuned, as in \cref{sec:setup}). $^\dagger$Sees the target; $^\P$removes or replaces fixtures; italic: the unedited input and the methods marked $^\dagger$ or $^\P$; bold and
underline: best and second among the other shared outputs.}
\label{tab:supp-lightlab}
\centering
\small
\setlength{\tabcolsep}{4pt}
\begin{tabular}{@{}lccc@{}}
\toprule
Method & PSNR$\uparrow$ & SSIM$\uparrow$ & LPIPS$\downarrow$ \\
\midrule
\multicolumn{4}{@{}l}{\emph{Outputs shared by LuxRemix's authors}} \\
LightLab~\cite{lightlab} & \textbf{19.25} & \underline{0.754} & \textbf{0.147} \\
LuxRemix, fitted to the target~\cite{liang2026luxremix} & \underline{18.75} & \textbf{0.781} & \underline{0.150} \\
ScribbleLight~\cite{scribblelight} & 15.56 & 0.504 & 0.457 \\
\textit{Qwen-Image}$^\P$~\cite{wu2025qwenimage} & \textit{20.81} & \textit{0.776} & \textit{0.148} \\
\midrule
\multicolumn{4}{@{}l}{\emph{Our runs}} \\
\textit{Unedited input} & \textit{18.46} & \textit{0.788} & \textit{0.144} \\
\textit{LumiNet}$^\dagger$~\cite{xing2025luminet} & \textit{18.19} & \textit{0.769} & \textit{0.198} \\
\textit{Qwen-Image-Edit}$^\P$~\cite{wu2025qwenimage} & \textit{20.12} & \textit{0.755} & \textit{0.167} \\
ScribbleLight~\cite{scribblelight} & 14.52 & 0.368 & 0.535 \\
Li \etal~\cite{li2022physically} & 17.08 & 0.660 & 0.472 \\
Ours & 15.84 & 0.613 & 0.500 \\
\bottomrule
\end{tabular}
\end{table}

LuxRemix~\cite{liang2026luxremix} evaluates real photographs on ten pairs of LightLab's evaluation set~\cite{lightlab}, the ten shown in the appendix of its paper (the full set of $200$ pairs is not released), with each method's edit fitted to the target (\cref{sec:comparisons}, \cref{tab:lux-lightlab}). It reports, on all ten (PSNR / SSIM / LPIPS), LightLab $21.80$ / $0.791$ / $0.144$, LuxRemix $20.93$ / $0.771$ / $0.168$, Qwen-Image $19.20$ / $0.735$ / $0.196$ and ScribbleLight $15.04$ / $0.465$ / $0.478$;
 its authors shared these pairs and their methods' saved outputs.
In six of the pairs one of the two photographs has no visible light on, so the room is lit only from outside the frame, and we
exclude them as on the paired photographs (\cref{sec:setup}).
On the remaining four, we score every method as LuxRemix does: each output resized to $512\times512$, whole-image PSNR, SSIM and
LPIPS in display space, without rescaling.
Each pair comes with LightLab's mask of the switched lights alone, which is all its model needs (\cref{fig:supp-theirmasks}).
We keep these masks, cutting one lamp's to its shade, and add the lights that stay lit and the windows with SAM~3 and the review of \cref{supp:masks}:
our model splits a photograph's shading among all its visible lights.
Ours and Li~\etal receive the same masks and are tuned as in \cref{sec:setup}, with one exposure for the whole image.
\Cref{fig:supp-lux-lightlab}a shows three of the pairs.
\Cref{tab:supp-lightlab} scores the outputs LuxRemix's authors shared and our own runs of every method, which \cref{tab:lux-lightlab}
reports (\cref{tab:lux-lightlab}).
The two ScribbleLight rows differ in their scribbles: ours are drawn from the light masks, as on the paired photographs, while LuxRemix does not state how the scribbles of the shared output were drawn; \cref{tab:lux-lightlab} compares our runs, made under one protocol.
LuxRemix's outputs keep the input photograph wherever their added light is dark: $53\%$ of their pixels are within $3/255$ of
the input, more than the real target ($20\%$), while every other method produces a new image.
No edit is closer to the target than the unedited input under LPIPS.
The methods trained on captured or rendered pairs of the same scene under different lights lead; ours trails them, as on the
rendered rooms of \cref{supp:luxremix}.
We edit at $256\times256$ and upsample to $512\times512$ for scoring; the unedited input resized the same way has LPIPS $0.306$
($0.144$ at full resolution).
Li \etal is ahead of ours on these four pairs, too few for the difference to be reliable.

\begin{figure*}[t]
  \centering
  \IfFileExists{figs/supp_controls.pdf}{\includegraphics[width=\linewidth]{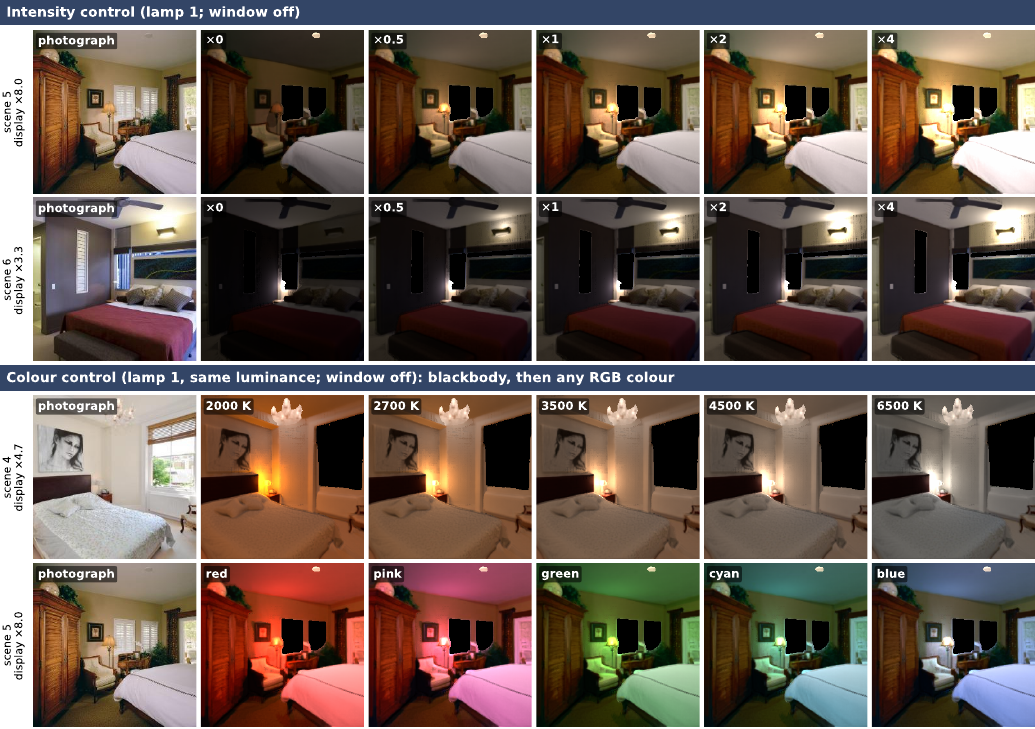}}{\fbox{\parbox[c][8cm][c]{0.9\linewidth}{\centering controls}}}
  \caption{\textbf{Editing each light on LSUN photographs.} Intensity: one lamp scaled $\times0$ to $\times4$. Colour: one lamp
  along the blackbody locus from $2000$ to $6500$\,K, and in arbitrary colours, at the same luminance. Every edited tile
  is albedo times the sum of the edited shading layers, with no pixel of the photograph; the exposure is fitted once per
  photograph, times one display gain per row (left).}
  \label{fig:supp-controls}
\end{figure*}
\begin{figure*}[p]
  \centering
  \includegraphics[width=0.86\linewidth]{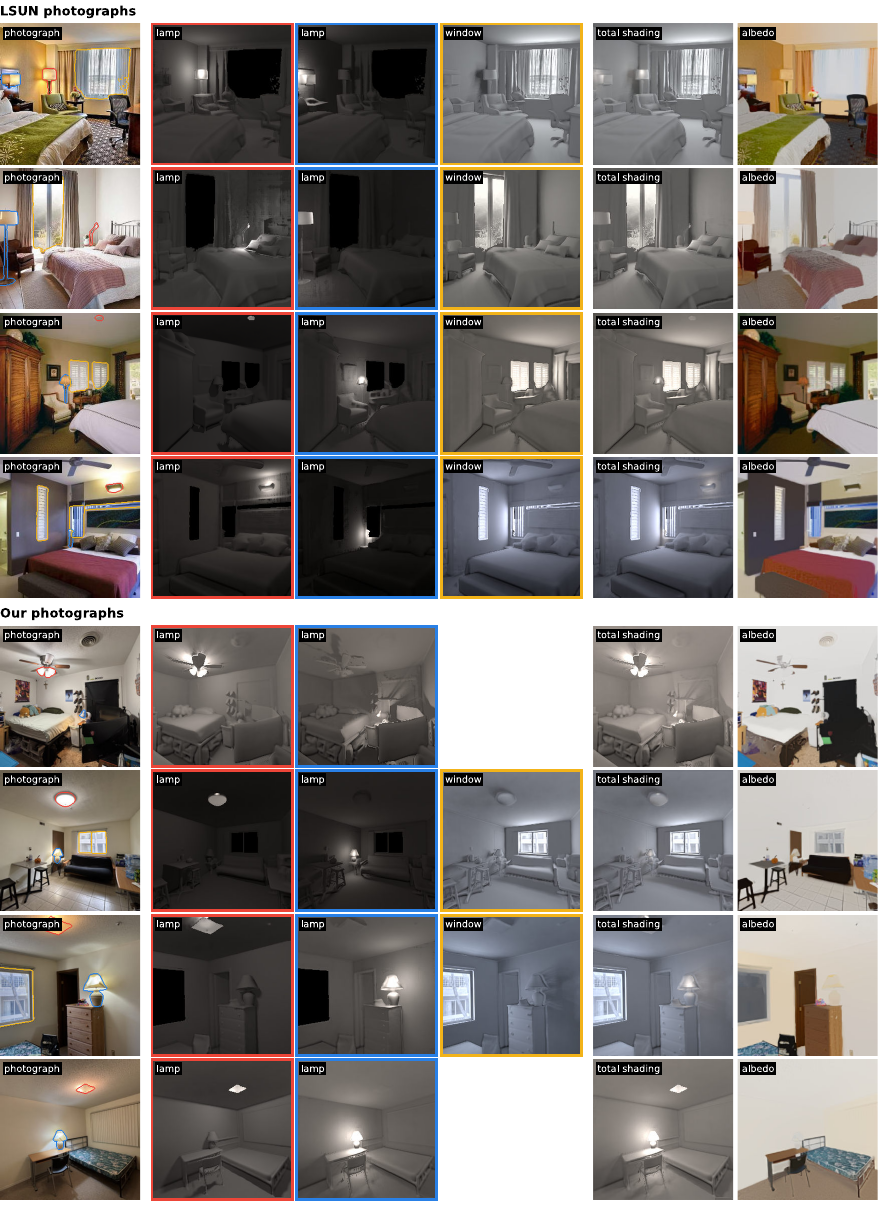}
  \caption{\textbf{More per-light shading layers.} Four LSUN photographs and four paired photographs, each with the lights that are on outlined
  in the colour of their layers' borders, the shading layer our model gives each of them (its emission, its direct light times its ray-traced visibility, and
  its indirect light), the total shading and the albedo; lamps include ceiling fixtures.}
  \label{fig:decomp-gallery}
\end{figure*}

\section*{More results}

\subsection{Editing each light}
\label{supp:controls}

Because every light has its own shading layer, an edit is a change to that light's parameters (\cref{sec:editing}, \cref{fig:teaser}): its intensity, its colour, or
switching it, for lamps and windows alike, and for several lights independently.
\Cref{fig:supp-controls} shows intensity and colour edits of a lamp on more photographs; the supplementary video also edits windows and several lamps in turn.
\Cref{fig:decomp-gallery} shows the per-light layers of eight photographs; as in \cref{fig:decomp-ours}, lamp layers are $\times3$ where a window lets in daylight.
Each edit takes
about $0.1$\,ms once the layers are computed (\cref{fig:cost}).

\paragraph{Supplementary video.}
The video (\texttt{supp\_video.mp4}, $51$\,s) shows edits on three LSUN photographs; each frame shows the photograph, the edited shading
and the edited image, the albedo times that shading.
0:03--0:06: one lamp brought from off to $\times3$, with the window off.
0:06--0:11: the lamp recoloured green, red and blue at equal perceived brightness.
0:11--0:17: the window turned on.
0:17--0:25: the window's light turned to sunset and then to a dimmer, overcast sky.
0:25--0:37: a second room's lamps switched and recoloured one at a time and its window faded out, the sequence of \cref{fig:teaser}.
0:37--0:51: direct sunlight cast through a window, swept in azimuth and then in elevation (below).

\paragraph{Direct sunlight.}
At inference, our model can cast direct sunlight through a window from an arbitrary azimuth and elevation, a light the photograph does not contain
(supplementary video, 0:37--0:51).
The sun is added at inference and is neither trained nor evaluated.
Its direction is an azimuth about the room's up axis, measured from the window's outward normal, and an elevation.
A pixel is sunlit if the ray from its 3D point towards the sun passes through the window's opening (the four-corner polygon of
the window mask, on the plane fitted to the wall around it, as for the window's visibility in \cref{supp:visibility}) without
hitting the depth mesh first.
The sun's direct light is its colour and intensity times the cosine between the surface normal and the sun direction, and its
bounce is the window's interreflected light scaled by the ratio of the sun's mean direct light to the window's.

\subsection{Shading quality, size and speed}
\label{supp:leakcost}
\input{tab_saw}
\begin{figure}[!htb]
  \centering
  \IfFileExists{figs/cost_scatter.pdf}{\includegraphics[width=\linewidth]{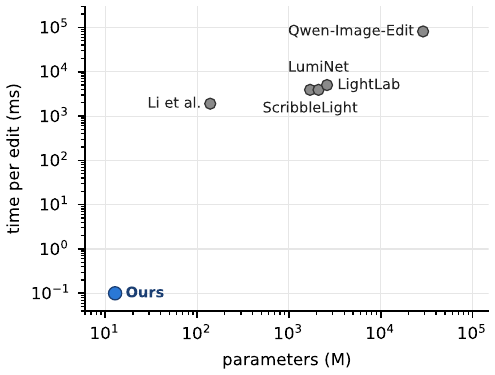}}{\fbox{\parbox[c][4cm][c]{0.9\linewidth}{\centering cost scatter}}}
  \caption{\textbf{Size and speed.} For the compared methods (\cref{tab:intro-methods}), parameters of the networks run for an edit
  against time per edit (log--log). Our model has $12.8$\,M parameters and takes about $0.1$\,ms per edit, after decomposing the photograph once
  ($95$\,ms for the network and about $2$\,s of ray casting on the CPU). The other methods with a reported time run again for every
  edit: Li \etal $1.9$\,s (as reported, without its optional optimisation), ScribbleLight and LumiNet $3.9$\,s, LightLab $5$\,s (as reported; backbone only) and
  Qwen-Image-Edit $81$\,s; LuxRemix ($\geq12$\,B) reports no time. Ours, ScribbleLight and LumiNet are timed on one NVIDIA A40, and Qwen-Image-Edit on two.}
  \label{fig:cost}
\end{figure}

Our shading keeps most of a photograph's reflectance out, scores within two points of the intrinsic model it learns from on SAW, and is edited in about $0.1$\,ms (\cref{sec:leak-cost}).
It keeps the reflectance out (\cref{fig:leak}) because its direct light and interreflection see the photograph only through the room code $\mathbf{z}$ and
each light's parameters; only the emission patterns, which are confined to the light-source masks, and the darkening map of the per-scene shading (\cref{supp:fallback}) are predicted from the photograph itself.
On SAW (\cref{tab:saw}), it reaches a challenge AP of $94.64$, against $96.48$ for Colorful Diffuse~\cite{careaga2024colorful}, whose shading it learns from, and it is the only shading in the table that is split per light.
The gap lies at shadow boundaries ($68.2$ against $90.4$), which we ray-cast against the estimated geometry rather than copy from the photograph; without the ray-traced shadows, our AP is $93.93$ and our shadow-boundary score $62.2$.
Our network has $12.8$\,M parameters and decomposes a photograph once; each edit then takes about $0.1$\,ms, against $1.9$ to $81$\,s for the other methods of \cref{fig:cost}, which run again for every edit.

\subsection{Failure cases}
\label{supp:failures}
\begin{figure*}[t]
  \centering
  \IfFileExists{figs/supp_failures.pdf}{\includegraphics[width=\linewidth]{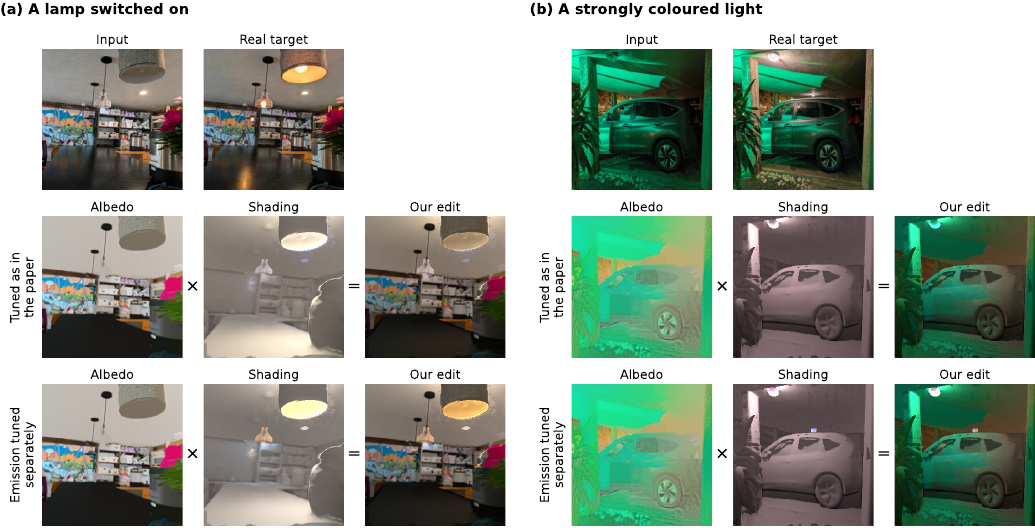}}{\fbox{\parbox[c][4cm][c]{0.9\linewidth}{\centering failures}}}
  \caption{\textbf{Failure cases on LightLab's pairs.} Each edit is its albedo times its shading. Middle row: our edit tuned as in the paper, with one gain per light.
  Bottom row: the same with a separate gain for each light's emission. (a)~The gain that suits the room leaves the shade pale; with its own gain, the shade glows as in the target.
  (b)~The green light is in the albedo and the shading is neutral, so neither tuning removes it.}
  \label{fig:supp-failures}
\end{figure*}

An edit re-renders the photograph as its albedo times the edited shading, so it is only as good as the albedo, which is computed once from the input photograph (\cref{sec:limitations}).
\Cref{fig:supp-failures} shows two cases: part of a strongly coloured light stays in the albedo (b), and our tuning scales a lamp's own glow and the light it casts with one gain (a), which tuning the glow separately removes.

Both are LightLab's real photograph pairs~\cite{lightlab}; (b) comes from the six pairs excluded from scoring (\cref{supp:lightlab}).
(a) A pendant lamp is switched on. Tuned as in the paper, one gain scales the lamp's emission and the light it casts together; the gain that suits the room leaves the shade pale, over an albedo taken with the lamp off. With a separate gain for the emission, the shade glows as in the target, and the rest of the edit is unchanged.
(b) Part of a strongly coloured light is baked into the albedo: the albedo is green and the shading neutral, so no tuning corrects the colours.

\begin{table}[t]
\caption{\textbf{User study: counts.} Choices in the comparisons of \cref{supp:userstudy}.}
\label{tab:supp-userstudy}
\centering
\footnotesize
\setlength{\tabcolsep}{4pt}
\begin{tabular}{@{}lrr@{}}
\toprule
Comparison & Chosen / shown & \% \\
\midrule
\multicolumn{3}{@{}l}{\emph{Against other methods}} \\
Ours over Li \etal & $219$ / $261$ & $84$ \\
Ours over ScribbleLight & $110$ / $118$ & $93$ \\
Ours over the real photograph & $7$ / $136$ & $5$ \\
Li \etal over the real photograph & $3$ / $134$ & $2$ \\
\midrule
\multicolumn{3}{@{}l}{\emph{Against the unedited photograph}} \\
Ours over the unedited photograph & $139$ / $151$ & $92$ \\
\bottomrule
\end{tabular}
\end{table}

\begin{figure*}[t]
  \centering
  \includegraphics[width=0.92\linewidth]{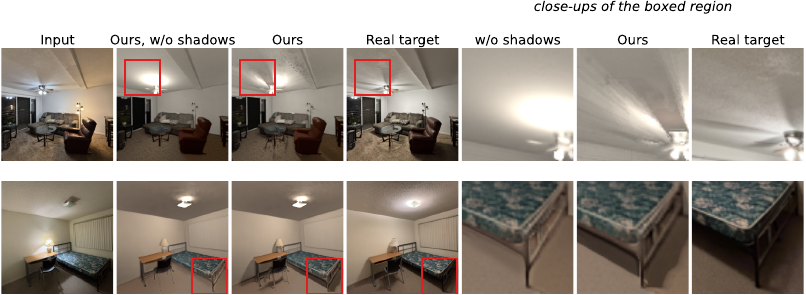}
  \caption{\textbf{Ray-traced shadows.} Our edit of two test pairs without and with the ray-traced visibility of \cref{supp:visibility}
  (the two variants of \cref{tab:ablation}), beside the real target. Top: with it, the fan's blades break up the ceiling fixture's light, as in the target; without it, the fixture lights the ceiling as a smooth blob.
  Bottom: with it, the bed shadows the floor beneath it, as in the target. Red box: the region of the close-ups.}
  \label{fig:supp-shadows}
\end{figure*}

\begin{table}[t]
\caption{\textbf{The paired photographs.} All taken with one phone on a tripod (\cref{supp:geometry}); a scene is one camera view of a room,
and a test pair is an ordered pair of its photographs whose labelled light states differ.
Scores are averaged over the pairs of a scene first and then over the $19$ scenes, so that scenes with many pairs do not dominate.}
\label{tab:supp-dataset}
\centering
\footnotesize
\setlength{\tabcolsep}{4pt}
\begin{tabular}{@{}lr@{}}
\toprule
Scenes (camera views) & $19$ \\
Switchable lights ($1$--$4$ per scene) & $40$ \\
Test pairs ($2$--$12$ per scene) & $128$ \\
\quad switching one / two / three or four lights & $80$ / $40$ / $8$ \\
\bottomrule
\end{tabular}
\end{table}

\begin{table}[t]
\caption{\textbf{Components.} The trained networks, $12.8$\,M parameters in all.}
\label{tab:supp-arch}
\centering
\footnotesize
\setlength{\tabcolsep}{3pt}
\begin{tabular}{@{}p{0.36\linewidth}p{0.44\linewidth}r@{}}
\toprule
Component & Output & Params \\
\midrule
\raggedright Extrinsic encoder, ResNet-18~\cite{he2016resnet} with RoI pooling & \raggedright room code $\mathbf{z}$, lamp features $\mathbf{h}_i$ & $11.3$\,M \\
\raggedright Interreflection-field CNN (\cref{supp:interrefl}) & \raggedright $a_{i}$, $\mathbf{d}_{i}$ of \cref{eq:field} & $1.1$\,M \\
\raggedright Other small CNNs and heads & \raggedright emission patterns $\hat\xi$, $\hat\psi$; per-scene shading; lamp parameters (\cref{eq:lamp-params}) & $0.4$\,M \\
\bottomrule
\end{tabular}
\end{table}

\begin{table}[t]
\caption{\textbf{Inside and outside the changed region.} PSNR against the target on the $128$ test pairs, scored as in \cref{sec:setup} and
averaged per scene, then over scenes. \emph{Changed}: the quarter of each image's pixels where the unedited photograph differs most
from the target; \emph{Rest}: the other pixels. Ours and Li \etal are tuned, as in \cref{tab:vs-unedited}. Ours is closer to the target than the unedited photograph inside the changed region on $87\%$ of the pairs. $^\dagger$Sees the target.}
\label{tab:supp-region}
\centering
\footnotesize
\setlength{\tabcolsep}{4pt}
\begin{tabular}{@{}lccc@{}}
\toprule
Method & Changed & Rest & Whole \\
\midrule
Unedited photo & 13.48 & 25.13 & 18.57 \\
\midrule
Ours & 16.72 & 19.87 & 18.66 \\
Li \etal & 16.37 & 20.67 & 18.99 \\
ScribbleLight & 14.60 & 16.68 & 15.93 \\
\midrule
Qwen-Image-Edit & 16.93 & 24.23 & 20.86 \\
LumiNet$^\dagger$ & 15.14 & 19.13 & 17.52 \\
\bottomrule
\end{tabular}
\end{table}

\begin{table*}[t]
\caption{\textbf{Training and inference cost}, collected from \cref{supp:s1train,supp:leakcost}. Training and all GPU times are on one NVIDIA A40; the network and edit times are medians at $256\times256$. The frozen pretrained models that prepare the inputs (masks, depth and normals, intrinsic decomposition) and the
albedo filter (\cref{supp:albedo}) also run once per photograph and are not included in the times.}
\label{tab:supp-cost}
\centering
\footnotesize
\setlength{\tabcolsep}{4pt}
\begin{tabular}{@{}ll@{}}
\toprule
\multicolumn{2}{@{}l}{\emph{Training}} \\
Data & $83$k LSUN bedrooms, $256\times256$ \\
Schedule & $1$ epoch, $5.2$k iterations, batch size $16$ \\
Optimiser & AdamW, weight decay $10^{-4}$ \\
Learning rate & $1.5\cdot10^{-4}$ \\
Training time & about $2$ GPU-hours \\
Hardware & one NVIDIA A40 \\
Trained parameters & $12.8$\,M \\
Framework & PyTorch \\
\midrule
\multicolumn{2}{@{}l}{\emph{Once per photograph}} \\
Network & $95$\,ms \\
Ray-cast visibility & about $2$\,s (CPU) \\
\midrule
\multicolumn{2}{@{}l}{\emph{Per edit}} \\
Re-weighting, to sRGB & $0.1$\,ms, $10$\,kHz \\
\bottomrule
\end{tabular}
\end{table*}

\begin{figure*}[t]
  \centering
  \includegraphics[width=\linewidth]{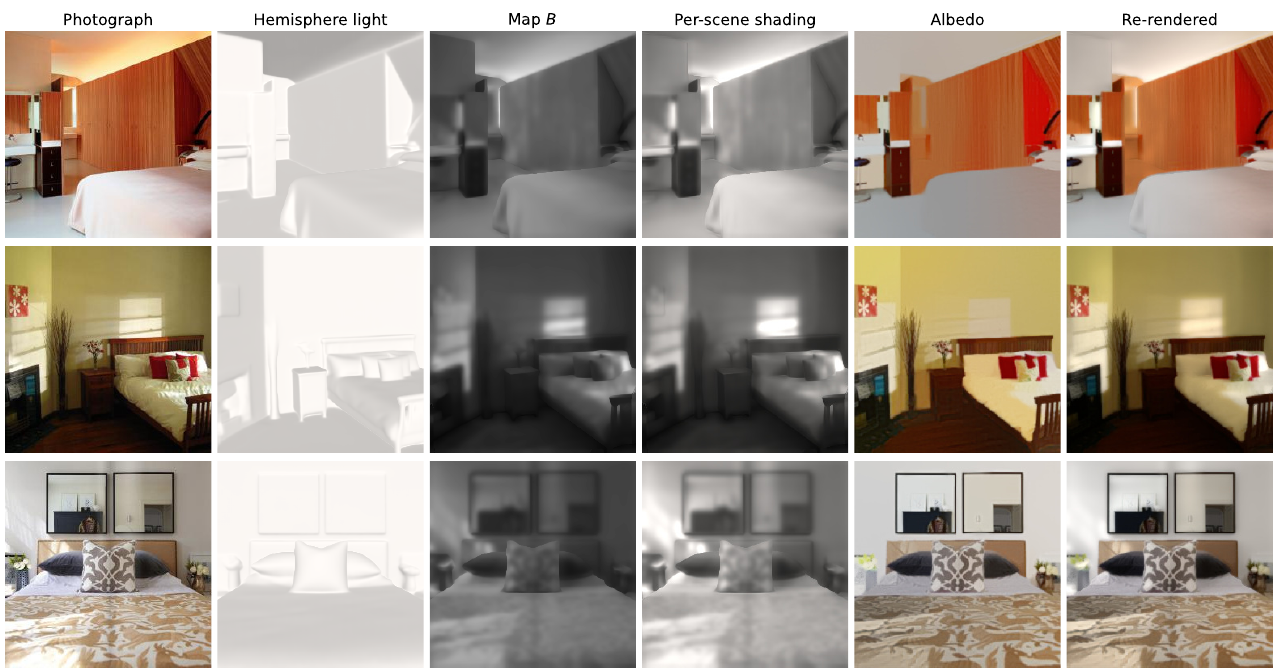}
  \caption{\textbf{Per-scene shading} (\cref{supp:fallback}) on three LSUN validation photographs with no detected light. The shading is the product of a hemisphere light,
  one colour and one direction per image, and the darkening map $B$, which gives it its spatial structure; the last column re-renders the photograph as albedo times this shading.
  $B$ is predicted from the photograph, so it also darkens some dark surfaces (the furniture, the picture frames); in the second row it carries the sunlight from a window outside the frame. Each shading is shown with its own display gain.}
  \label{fig:supp-fallback}
\end{figure*}

\clearpage

\clearpage

%% file: tab_saw.tex
\providecommand{\hcell}[1]{\begin{tabular}[b]{@{}c@{}}#1\end{tabular}}
\begin{table}[t]
\caption{\textbf{Shading on SAW.} Challenge AP~\cite{li2018cgintrinsics} of the total shading on the SAW test
photographs~\cite{kovacs2017saw}: all $1{,}699$, and the $1{,}219$ with a detected lamp, ceiling fixture or window.
Shadow bound.: the share of shadow-boundary points classified correctly (all photographs) at the threshold that
classifies $90\%$ of smooth-shading points correctly. We score every row with one scorer; it reproduces the published
$97.93$ of CGIntrinsics, and Ordinal Shading~\cite{careaga2023ordinal} reports $95.5$. Li \etal~\cite{li2022physically} is
not scored: with the optimisation of its light parameters, it took several minutes per photograph in our runs (its reported $1.9$\,s per edit is without this optimisation), too slow for $1{,}699$ photographs.
Bold, underlined: best, second. $^*$Trained on SAW's training annotations. $^\dagger$The intrinsic model whose shading we learn from.}
\label{tab:saw}
\centering
\small
\setlength{\tabcolsep}{4.5pt}
\begin{tabular}{@{}lcccc@{}}
\toprule
 & & \multicolumn{2}{c}{SAW AP$\uparrow$} & \\
\cmidrule(lr){3-4}
Method & Per light & All & \hcell{With\\lights} & \hcell{Shadow\\bound.$\uparrow$} \\
\midrule
CGIntrinsics~\cite{li2018cgintrinsics}$^*$          & --         & \underline{97.93} & \underline{97.76} & \underline{91.8}  \\
Ordinal Shading~\cite{careaga2023ordinal}           & --         & 94.70             & 94.76             & 77.7              \\
Colorful Diffuse~\cite{careaga2024colorful}$^\dagger$ & --       & 96.48             & 96.61             & 90.4              \\
IntrinsicDiffusion~\cite{luo2024intrinsicdiffusion} & --         & \textbf{98.48}    & \textbf{98.65}    & \textbf{97.8}     \\
\midrule
Ours, no shadows                                    & \checkmark & 93.93             & 94.02             & 62.2              \\
Ours                                                & \checkmark & 94.64             & 94.76             & 68.2              \\
\bottomrule
\end{tabular}

\end{table}